\documentclass[journal]{IEEEtran}

\IEEEoverridecommandlockouts                              %
\makeatletter

\let\proof\@undefined
\let\endproof\@undefined
\makeatother

\usepackage{amsmath} %
\usepackage{amssymb}  %
\usepackage{amsthm}
\usepackage{amsfonts}
\usepackage{mathtools}

\newtheorem*{problem*}{Problem}
\newtheorem{nono-problem}{Problem}[]

\theoremstyle{definition}

\theoremstyle{remark}

\usepackage{verbatim}

\usepackage{color}
\usepackage{subfig}
\usepackage{graphicx}
\usepackage{caption}
\graphicspath{{images/}}

\usepackage{mathptmx}
\usepackage{times}

\usepackage[ruled,vlined,linesnumbered]{algorithm2e}
\usepackage[noend]{algorithmic}

\usepackage[us]{datetime}
\usepackage{caption}
\newcommand\red[1]{{\textcolor{red}{#1}}}

\usepackage[usenames,dvipsnames,table]{xcolor}

\usepackage{hyperref} %
\hypersetup{
    colorlinks,
    citecolor=black,
    filecolor=black,
    linkcolor=black,
    urlcolor=black,
    pdfauthor={},
    pdfsubject={},
    pdftitle={}
}

\usepackage{todonotes}

\usepackage{graphicx}

\usepackage{subfig}

\usepackage{latexsym}
\usepackage{color}

\usepackage{cite}

\usepackage{caption}
\usepackage{booktabs} %
\usepackage{bm}
\usepackage{multirow}

\newcommand\oprocendsymbol{\hbox{$\square$}}
\newcommand\oprocend{\relax\ifmmode\else\unskip\hfill\fi\oprocendsymbol}

\newcommand{\hiddencomment}[2]{\red{#1}}
\renewcommand{\hiddencomment}[2]{}

\usepackage{paralist}

\makeatletter
\renewcommand{\dddot}[1]{%
  {\mathop{\kern\z@#1}\limits^{\vbox to-1.4\ex@{\kern-\tw@\ex@
   \hbox{\normalfont ...}\vss}}}}
\renewcommand{\ddddot}[1]{%
  {\mathop{\kern\z@#1}\limits^{\vbox to-1.4\ex@{\kern-\tw@\ex@
   \hbox{\normalfont....}\vss}}}}
\makeatother
\usepackage{bm}
\makeatletter 
 \def\@eqnnum{{\normalsize \normalcolor (\theequation)}} 
  \makeatother

\newcommand{\mass}{m}				%
\newcommand{\massR}{\mass_R}			%
\newcommand{\offsetCoM}{\mathbf{r}_{BL}}	%

\newcommand{\inertia}{J}			%
\newcommand{\inertiaR}{\inertia_R}		%

\newcommand{\pitch}{\vartheta}			%
\newcommand{\dpitch}{\dot{\pitch}}		%
\newcommand{\ddpitch}{\ddot{\pitch}}		%
\newcommand{\elevation}{\varphi}		%
\newcommand{\delevation}{\dot{\elevation}}	%
\newcommand{\ddelevation}{\ddot{\elevation}}	%
\newcommand{\unitv}{\mathbf{d}}			%

\newcommand{\state}{\mathbf{x}}			%
\newcommand{\dstate}{\dot{\state}}		%
\newcommand{\inp}{\mathbf{u}}			%
\renewcommand{\output}{\mathbf{y}}		%
\newcommand{\modCon}{a}				%

\newcommand{\genCoordinates}{\mathbf{q}}	%

\newcommand{\thrust}{f_R}			%
\newcommand{\dthrust}{\dot{f}_R}		%
\newcommand{\bthrust}{\mathbf{f}_R}		%
\newcommand{\torque}{\tau_R}			%

\newcommand{\stress}{f_L}			%

\newcommand{\bx}{\mathbf{x}}			%
\newcommand{\by}{\mathbf{y}}			%
\newcommand{\bz}{\mathbf{z}}			%
\newcommand{\bu}{\mathbf{u}}
\newcommand{\bh}{\mathbf{h}}
\newcommand{\disturbance}{\boldsymbol{\delta}}	%

\newcommand{\bzeta}{{\boldsymbol{\zeta}}}	%
\newcommand{\inObs}{\mu}			%
\newcommand{\bInObs}{{\boldsymbol{\inObs}}}	%

\newcommand{\decouplingM}{\mathbf{E}}		%
\newcommand{\virtualInput}{\mathbf{v}}		%

\newcommand{\ddposR}{\ddot{\mathbf{p}}_B}	%
\newcommand{\accMeasures}{\mathbf{a}}		%
\newcommand{\measure}{w}			%

\newcommand{\quality}{\tilde{e}}		%

\newcommand{\wFrame}{\mathcal{F}_W}		%
\newcommand{\bFrame}{\mathcal{F}_B}		%
\newcommand{\lFrame}{\mathcal{F}_L}		%

\newcommand{\nR}[1]{\mathbb{R}^{#1}}		%
\newcommand{\norm}[1]{\left\lVert#1\right\rVert} %
\newcommand{\matrice}[1]{\begin{bmatrix} #1 \end{bmatrix}}	%
\newcommand{\cosp}[1]{\cos \left( #1 \right)}	%
\newcommand{\sinp}[1]{\sin \left( #1 \right)}	%
\newcommand{\cosS}[1]{c_{#1}}		%
\newcommand{\sinS}[1]{s_{#1}}		%
\newcommand{\upperRomannumeral}[1]{\uppercase\expandafter{\romannumeral#1}}	%

\newcommand{\fig}{Fig.~}	%
\newcommand{\eqn}[1]{(#1)}	%
\newcommand{\tab}{Tab.~}	%
\newcommand{\sect}{Sec.~}	%

\title{Dynamics,  Control, and Estimation for Aerial Robots Tethered by Cables or  Bars}

\author{Marco Tognon$^{1}$ and Antonio Franchi$^{1}$
\thanks{$^1$LAAS-CNRS, Universit\'e de Toulouse, CNRS, Toulouse, France,   \href{mailto:marco.tognon@laas.fr}{marco.tognon@laas.fr}, \href{mailto:antonio.franchi@laas.fr}{antonio.franchi@laas.fr}}
\thanks{
This work has been partially funded by the European Union's Horizon 2020 research and innovation programme under grant agreement No 644271 AEROARMS.}
}

\makeatletter
\def\ps@titlepagestyle{
	\def\@oddfoot{}\def\@evenfoot{}
	\def\@oddhead{\textcolor{red}{\sf }}
	\def\@evenhead{\textcolor{red}{\sf }}%
}%
\makeatother
\makeatletter
\def\ps@headings{
	\def\@oddfoot{\textcolor{red}{\small\sf  Preprint version final version at http://ieeexplore.ieee.org/ \hfill \thepage \hfill Accepted for IEEE Transaction on Robotics 2016}}\def\@evenfoot{\hfill\thepage\hfill}
	\def\@oddhead{}\def\@evenhead{}%
}%
\makeatother
    \makeatletter
    \def\ps@IEEEtitlepagestyle{%
      \def\@oddhead{\small \textcolor{red}{\sf  Preprint version final version at http://ieeexplore.ieee.org/ \hfill  Accepted for IEEE Transaction on Robotics 2017}}
      \def\@evenhead{}%
      \def\@oddfoot{\hfill \thepage \hfill}%
      \def\@evenfoot{}
    }
    \makeatother
\begin{document}

\maketitle

\begin{abstract}
We consider the problem of controlling an aerial robot connected to the ground by a passive cable or a passive rigid link. 
We provide a thorough characterization of this nonlinear dynamical robotic system in terms of fundamental properties such as differential flatness, controllability, and observability. 
We prove that the robotic  system is differentially flat with respect to two output pairs: elevation of the link and attitude of the vehicle; elevation of the link  and longitudinal link force (e.g., cable tension, or bar compression).
We show the design of an almost globally convergent nonlinear observer of the full state that resorts only to an onboard accelerometer and a gyroscope. 
We also design two almost globally convergent nonlinear controllers to track any sufficiently smooth time-varying trajectory of the two output pairs.
Finally we numerically test the robustness of the proposed method in several far-from-nominal conditions: nonlinear cross-coupling effects, parameter deviations, measurements noise and non ideal actuators.
\end{abstract}
\section{Introduction}\label{sec:intro}

Research on aerial robots has been boosted in the last decade thanks to the availability of low-cost and light-weight technology for onboard sensing and computation.
This increase of popularity led to development of several methods for different problems  such as, e.g., flight control, aerial navigation, planning, sensing, mapping, and transportation.
Although the main use of aerial robots has been the one of mobile remote sensors, today their use for physically interactive tasks has become a very popular and promising topic, like the cooperative transportation of suspended loads with cables~\cite{2010-MazKonBerOll,2013-ManDevRosCor}.
Cables are not only used for cooperative transportation but also to tether an aerial vehicle to a ground station. 
In this case the cable can provide many advantages to the robotic system, such as\begin{inparaenum}[\em i)] 
\item improved flight stability and reliability 
\item physical interaction with a ground object,
\item enduring power, and
\item high bandwidth communication channel with the base station.
\end{inparaenum}
Applications fields related to this kind of robotics systems are 
\begin{inparaenum}[\em i)]
\item landing/taking-off from/on moving or sloped platforms~\cite{2016c-TogDasFra,2016i-TogTesRosFra,2006-Oh_PatAgrPotGar,2013-SanBejKonOll},
\item inspection and surveillance~\cite{2011-Mut}, and
\item communication reinforcing using relays.%
\end{inparaenum}

For an aerial vehicle connected by a passive\footnote{The `passive' adjective  differentiates from the `active' case in which the link is attached to a controlled winch that can regulate its length.} link to the ground, standard flight-control and estimation methods cannot be applied straightforwardly because they do not exploit the full dynamics and capabilities of the new interconnected system.  
However, it is hard to design new control and estimation methods that consider the system in its whole completeness, due to the nonlinear dynamics and the dynamic coupling between the aerial vehicle and the link.

Driven by the relevance of this topic, several control and estimation schemes have been presented in the robotic literature. 
For the case in which the aerial vehicle is a helicopter the authors of~\cite{2006-Oh_PatAgrPotGar} present a method to land it on the deck of a ship in rough sea using a cable. The controller is based on a time-scale separation technique between the rotational and translational dynamics.
In the context of~\cite{ec-safemobil}, the authors of~\cite{2013-SanBejKonOll} present a control scheme based on a PID scheme together with partial model inversion, to stabilize the flight to a constant horizontal elevation with the goal of both improving hovering in windy conditions and land on a mobile platform. 

For the case of cable-tethered underactuated multi-rotor system,~\cite{2013-LupDan} presents a controller that, under a quasi-static assumption, stabilizes the elevation (angle) of the cable to a constant value using only inertial onboard sensors. 
The authors of~\cite{2014-NicNalGar} also present a controller to stabilize the elevation of a tethered multi-rotor to a constant value but they ensure the positivity of the cable tension as well.

The presented approaches provide a good basis for observation and control  of such systems. However, to improve those methods and overcome some of their drawbacks we want to:

\begin{inparaenum}
	\item consider a more \textit{general link} (i.e., a cable or a bar), differently from the state-of-the-art where only a cable link is used. Indeed the use of rigid link can be beneficial for, e.g., sustain part of the platform weight, or allow a full-bilateral physical interaction with a ground object.
	
	\item \textit{track a time varying trajectory}. While the methods in the literature are based on a quasi-static assumption, in many practical cases it is important to track a time-varying trajectory to, e.g., let the aerial robot track a moving target.
	
	\item \textit{track a desired link force profile}. In many cases is important to precisely regulate the force link to any desired value, possibly including negative link force (pushing) if the link is a bar.
	 However, the methods in the literature are able, in the best case~\cite{2014-NicNalGar}, to keep the force link generically positive.
	
	\item \textit{observe the state} from onboard IMU measurements in any dynamic conditions, while in the state-of-the-art this is done at best in a quasi static assumption~\cite{2013-LupDan}.
\end{inparaenum}	

In order to achieve these goals we started by analyzing in details the intrinsic properties of the system such as differential flatness, controllability and observability. Furthermore, we address three main control objectives, described in the following.

The first objective is to solve the tracking problem for any time-varying desired link-elevation trajectory while, at the same time, letting the aerial vehicle attitude track any other time-varying trajectory as well. This objective responds to applications requiring to place the robot at a desired altitude and with a desired orientation, e.g., for surveillance-task.

The second objective is to solve the elevation tracking problem together with the problem of tracking  time-varying force-link trajectories that can assume values both in the positive (tensions) and negative (compressions) domains. For example, if the link is a cable one may like to keep it always taut while at the same time avoid peaks on the tension that can damage the cable or its attaching mechanisms. If the link is a bar then one may want to push against the bar appropriately to perform physical interaction with the environment.

The third and last objective is to close the control loop with a minimal sensorial setup. For this purpose we design a nonlinear observer to retrieve the state from any dynamic case using inertial-only onboard measurements, without the need of GPS or vision sensors. 
This makes the system feasible for both inside and outside environments and also in the case of very low or zero visibility. 
The ability to work with a minimum number of sensors also increases the robustness to the failure of the other sensors without precluding their additional use.

The paper is organized as follows. 
In \sect\ref{sec:model} we derive the dynamic model and we describe the addressed control problems.
In the following \sect\ref{sec:flatness} we prove the differential flatness of the system with respect to the two sets of outputs. 
Then, in \sect\ref{sec:control} we design the controllers for the tracking of these outputs.
The observability analysis and the derivation of the observer are presented in \sect\ref{sec:observer}.
A comprehensive numerical validation of the methods is shown in \sect\ref{sec:simulations}.
Finally, in \sect\ref{sec:conclusion}, conclusions and future works are discussed.

\section{Dynamic Model and Control Problem}\label{sec:model}

\begin{figure}[t]
	\centering
	\includegraphics[height=0.4\columnwidth]{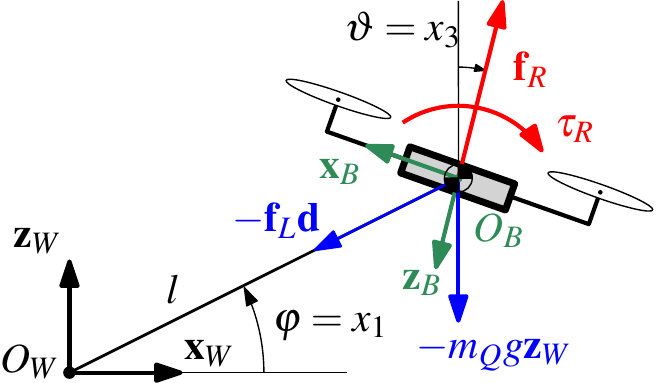}
\caption{Schematic representation of the  system for which two controllers and one observer are designed for the control of the two output pairs: $(\elevation,\pitch)$, and $(\elevation,\stress)$. The presence of two opposing propellers is just illustrative, since the same inputs can be realized with different solutions as, e.g., a ducted fan configuration, or another multi-propeller configuration.}
	\label{fig:model:system_model}
\end{figure}

Similarly to~\cite{2013-LupDan,2014-NicNalGar} we consider a robotic system composed by  an underactuated \emph{aerial vehicle} moving on a vertical plane and tethered by a \textit{link} to the ground, as depicted in~\fig\ref{fig:model:system_model}. 
We define 
an inertial \textit{world frame}, $\wFrame$, with axes $\{ \bx_W, \by_W, \bz_W\}$ and origin $O_W$, 
and a \textit{body frame}, $\bFrame$, attached to the aerial vehicle, with axes $\{ \bx_B, \by_B, \bz_B\}$ and origin $O_B$ coinciding with the center of mass (CoM) of the aerial vehicle. 
The position of $O_B$,  defined in $\wFrame$, is denoted with $\mathbf{p}_B = \left[x_B\;\, y_B\;\, z_B\right]^T \in\mathbb{R}^3$.
Since the vehicle moves on the vertical plane, we have that $\by_B\equiv\by_W$ and $y_B\equiv0$.\footnote{Although the system evolves in a 2D plane, defining the frame in 3D is convenient in order, e.g., to define properly the torque and angular velocity vectors for the robot.} 
The orientation of $\bFrame$ is fully known once it is known the angle of rotation $\pitch$  needed to let the orientation of $\wFrame$ coincide with the one of $\bFrame$ when performing a rotation about $\by_W\equiv\by_B$.

The aerial vehicle is modeled as a rigid body of mass $\massR\in\nR{}_{>0}$ and rotational inertia $\inertiaR\in\nR{}_{>0}$. 
Its (untethered) configuration is fully described by $x_B$, $z_B$, and the \emph{attitude} $\pitch$. 
The vehicle is actuated by two scalar inputs: 
\begin{inparaenum}[\em i)]
\item the intensity, $\thrust\in\nR{}$, of the thrust force vector $ \bthrust=-\thrust\bz_B$, and 
\item the intensity  $\torque\in\nR{}$, of the torque vector $\torque\by_B$.
\end{inparaenum}
The aerial vehicle is equipped with an inertial sensor composed by 
\begin{inparaenum}[\em i)]
\item a two-axis accelerometer with axes aligned with $\bx_B$ and $\bz_B$  and 
\item a one-axis gyroscope aligned with $\by_B$,
\end{inparaenum}
 whose measurements are denoted by $\accMeasures = [a_x\;a_z]^T\in\nR{2}$
and $\omega\in\nR{}$, respectively, where $\omega\by_B$ is the angular velocity of the aerial vehicle.

The link 
can be either a bar, a strut or a cable. It is connected  at one end to $O_B$ and at the other end to the fixed point   $O_W$. Connections are modeled as two passive rotational joints, therefore no rotational constraint holds at the joints. 
As done in related works, see, e.g.~\cite{2012-NalGasGar,2015a-TogFra,2013-LupDan,2014-NicNalGar}, we assume that the mass and inertia of the link are negligible with respect to the inertial characteristic of the aerial vehicle. However, this assumption is practically relaxed in \sect\ref{sec:simulations} where we test the proposed method with non-zero mass and inertia of the link. Deformations and elasticity are also negligible in the operative condition.
Therefore the link length, denoted with $l\in\nR{}_{>0}$, is fixed.  The  configuration of the link is described by the \textit{elevation}, $\elevation\in\nR{}$, i.e., the angle that the link forms with the $\bx_W$ axis. %
We denote with $\stress\in\nR{}$ the intensity of the internal force acting along the link, called \textit{link force}. When the link is pulled the force is called \textit{tension} and $\stress > 0$, whereas when it is compressed the link force is called \textit{compression} and $\stress < 0$.

The mechanical model of the robotic system (i.e., aerial vehicle plus link) can be then easily derived. Given the constraints of the system, the motion is restricted on a circle of radius $l$. 
Thus the configuration of the system is fully described by the generalized coordinates $\genCoordinates = [\elevation\;\,\pitch]^T$. Applying the Euler-Lagrange approach it is possible to  obtain the equations of  the dynamic model
\begin{align}
\mathbf{M}(\genCoordinates)\ddot{\genCoordinates} + \mathbf{g}(\genCoordinates) = \mathbf{Q}(\genCoordinates)\inp
\label{eqn:model:lagrangianModel}
\end{align}
where
\begin{gather*}
\mathbf{M} = 
\matrice{\inertia_\elevation	&	0 \\
		 0	&	\inertia_\pitch	} =
\matrice{\massR l^2 & 0 \\ 0 & \inertiaR}		 , \quad
\mathbf{g} = \matrice{\mass_R l g \unitv^\perp \cdot \mathbf{e}_3 \\ 
		0 }, \\
\mathbf{Q} = \matrice{-l \mathbf{R}_B^W\mathbf{e}_3\cdot \unitv^\perp	&	0	\\
         0 & 1 },\quad
         \inp = \matrice{ \thrust \\ \torque }=\matrice{ u_1 \\ u_2 },			 
\end{gather*}%
and:\footnote{As normal in the literature
 we define $\sinS{\star} = \sinp{\star}$ and $\cosS{\star} = \cosp{\star}$.} $\unitv = [\cosS{\elevation} \; 0 \;\sinS{\elevation}]^T$, 
$\unitv^\perp=[-\sinS{\elevation} \; 0 \;\cosS{\elevation}]^T$
$\mathbf{R}_B^W\in SO(3)$ is the rotation matrix from $\bFrame$ to $\wFrame$, $g\approx 9.81$, and\footnote{$\mathbf{e}_i\in\nR{3}$ is the canonical vector with $1$ in position $i-th$ and zero otherwise.} $\mathbf{e}_3 = [0 \;\,0 \;\,1]^T$.

Defining the state of the system by $\state = \left[ \elevation\;\, \delevation\;\, \pitch\;\, \dpitch \right]^T=\left[ x_1\;\, x_2\;\, x_3\;\, x_4 \right]^T$, the dynamic model in state space form results:
\begin{align}\label{eqn:model}
\dstate = \matrice{x_2 \\ \modCon_1\cosS{x_1} \\ x_4 \\ 0} + 
\matrice{0	& 0 \\
		 \modCon_2\cosS{x_1 + x_3} & 0 \\
		 0 & 0 \\
		 0 & \modCon_3}\inp, 
\end{align}
where $\modCon_1 = -g/l$, $\modCon_2 = 1/(\massR l)$, $\modCon_3 = 1/\inertiaR$ are the constant parameters of the dynamical model.

The problems addressed in this paper are the following.
\begin{problem*}[Addressed Control Problems]
Design two control laws for the input $\inp$, computing the control action using only the onboard measurements $\accMeasures$ and $\omega$ and each achieving the following two different control objectives, respectively.

\emph{Objective 1}:  let $\elevation$ and $\pitch$ independently track the desired time-varying trajectories $\elevation^d(t)$ and $\pitch^d(t)$, respectively;

\emph{Objective 2}: let $\elevation$ and $\stress$ independently track the  desired time-varying trajectories $\elevation^d(t)$ and $\stress^d(t)$, respectively. 

\end{problem*}

In view of the given control objectives we define the two pairs of outputs: $\output^a = \left[\elevation\;\, \pitch \right]^T = \left[ y^a_{1}\;\, y^a_{2}\right]^T$ and $\output^b = \left[\elevation\;\, \stress \right]^T = \left[ y^b_{1}\;\, y^b_{2}\right]^T$.
In order to write the outputs as a function of $\state$ and $\inp$ we first retrieve the link force from the force balance equation
\begin{align}
	\massR \ddposR = \massR(-l\mathbf{d}\delevation^2 + l\mathbf{d}^\perp\ddelevation)= - \stress \unitv - \thrust \bz_B - \massR g\bz_W,	\label{eqn:model:balanceForces}
\end{align}
from which we obtain:
\begin{align}
\output^a &= \matrice{x_1 & x_3}^T \label{eqn:model:output:a}\\
\output^b &= \matrice{	x_1\\ \frac{1}{\modCon_2} x_2^2 + \frac{\modCon_1}{\modCon_2}\sinS{x_1}} + \matrice{0 & 0\\ \sinS{x_1 + x_3} & 0 }\inp. \label{eqn:model:output:b}
\end{align}

\section{Differential Flatness}\label{sec:flatness}

Differential flatness is a property of certain dynamical systems  for which it exists (at least) an output (with the same dimension of the input), called \textit{flat output}, such that both the state and the input can be expressed as an algebraic function of the flat output and its derivatives, up to a finite order~\cite{1995-FliLevMarRou}. 
This property is commonly used to generate feasible trajectories for the state and the input by planning only in the output space, which is useful for nonholonomic and underactuated systems.

In the following we demonstrate that the system at hand is differentially flat with respect to both outputs, i.e., $\state$ and $\inp$ can be written as functions of either $\output_a$ or $\output_b$ and their derivatives.

\subsection{Flatness with Respect to $\output^a$}\label{subsec:flatness:a}

From the definition of $\output^a$ we have directly that $\state = \matrice{y^a_1  & \dot{y}^a_1 & y^a_2 & \dot{y}^a_2}^T$.
Then, inverting~\eqref{eqn:model} we can write the inputs as a function of the output and some of its derivatives
\begin{align*}
u_1 &= \frac{\dot x_2 - \modCon_1\cosS{x_1}}{\modCon_2 \cosS{x_1 + x_3}} = \frac{\ddot y^a_1 - \modCon_1\cosS{y^a_1}}{\modCon_2 \cosS{y^a_1 + y^a_2}} = u_1(y^a_1,\dot y^a_1, \ddot y^a_1,y^a_2) \\
u_2 &= {\dot x_4}/{\modCon_3} = {\ddot y^a_2}/{\modCon_3} = u_2(\ddot y^a_2). 
\end{align*}
Notice that in order to compute $u_1$ as above, it has to hold that $y_1^a + y_2^a = \elevation + \pitch \neq \pi/2 + k\pi$ where $k\in\mathbb{Z}$.
In the output plane, the singularity represents a family of parallel lines. This means that if initial and final outputs configurations are on the opposite side of one line, then there are no trajectories connecting the two configurations that avoid the singularity in which $u_1$ results undetermined. The same singularity will appear in the controller and in \sect\ref{subsec:controllers:ab} we address the issue and we propose a possible solution to it.

\subsection{Flatness with Respect to  $\output^b$}\label{subsec:flatness:b}

First of all we have directly $x_1 = \elevation = y^b_1$ and $x_2 = \delevation = \dot{y}^b_1$.
Then consider $\mathbf{r} = -\massR \ddposR(y^b_1,\dot y^b_1,\ddot y^b_1,y^b_2) - y^b_2 \mathbf{d}(y^b_1) -\massR g\bz_W$ that depends only on $y^b_1,\dot y^b_1, \ddot y^b_1,y^b_2$ and some constant parameters of the system. Using~\eqn{\ref{eqn:model:balanceForces}} we get
\begin{align}
u_1 \bz_B = u_1 [-\sinS{x_3}\;\, 0 \,\;-\cosS{x_3}]^T = \mathbf{r}(y^b_1,\dot y^b_1, \ddot y^b_1,y^b_2),\label{eqn:model:balanceForces_r}
\end{align}
where $\mathbf{r}(\cdot,\cdot,\cdot,\cdot) =  [r_1 \;\, r_2 \;\, r_3 ]^T$ represents the function  from ${\mathbb{R}}^2 \times {\mathbb{R}}^2 \times {\mathbb{R}}^2 \times {\mathbb{R}}^2$ to ${\mathbb{R}}^3$.
If $\norm{\mathbf{r}} \neq 0$, from \eqref{eqn:model:balanceForces_r} we can retrieve $u_1$ and $x_3$ as
\begin{align}\label{eqm:flatness:posSol}
\begin{split}
u_1 &= 	\norm{\mathbf{r}} = u_1(y^b_1,\dot y^b_1, \ddot y^b_1,y^b_2) \\
x_3 &= \textrm{atan2}\left({r_1}/{\|\mathbf{r}\|},{r_3}/{\|\mathbf{r}\|}\right) = x_3 (y^b_1,\dot y^b_1, \ddot y^b_1,y^b_2).
\end{split}
\end{align}
Differentiating $x_3$ w.r.t. time we  obtain $x_4$
\begin{align*}
x_4 &= \frac{d}{dt} x_3 (y^b_1,\dot y^b_1, \ddot y^b_1,y^b_2)= x_4 (y^b_1,\dot y^b_1, \ddot y^b_1, {y_1^b}^{(3)},y^b_2,\dot y^b_2). 
\end{align*}
Further differentiating $x_4$  and using~\eqref{eqn:model} we finally obtain $u_2$
\begin{align*}
u_2 &= \inertiaR \frac{d}{dt} x_4 (y^b_1,\ldots, {y_1^b}^{(3)},y^b_2,\dot y^b_2)= u_2 (y^b_1,\ldots,{y_1^b}^{(4)},y^b_2,\dot y^b_2,\ddot y^b_2),
\end{align*}
which concludes the proof that $\output^b$ is a flat output for the system in exam in the case $\|\mathbf{r}\| \neq 0$.
An extended version of the previous equations can be found in \cite{2017-TogFra}.

\subsubsection{Case of Vanishing Thrust}\label{subsec:vanishingThrust}

If $\|\mathbf{r}\| = 0$ one can still obtain the state and the inputs in terms of the outputs. Differentiating both sides of~\eqref{eqn:model:balanceForces_r}
and substituting $u_1 = \|\mathbf{r}\| = 0$ we obtain 
$\dot{u}_1 \bz_B = \dot{\mathbf{r}}(y^b_1,\dot y^b_1, \ddot y^b_1, {y_1^b}^{(3)},y^b_2,\dot y^b_2)$
from which, if $\|\dot{\mathbf{r}}\|\neq 0$, we can compute $\dot u_1$ and $x_3,x_4,u_2$ as before.
Similarly, if $\norm{\mathbf{r}}=\norm{\dot{\mathbf{r}}} =\ldots= \norm{\mathbf{r}^{(r-1)}}=0$ and $\norm{\mathbf{r}^{r}}\neq0$, we can compute the inputs and the state differentiating \eqref{eqn:model:balanceForces_r} $r$ times and then applying the same method as before.

If all the derivatives of $u_1=\|\mathbf{r}\|$ are zero along the desired output trajectory then $x_3,x_4$ and $u_2$ are indefinite.  In this degenerate case, the thrust is identically zero, $u_1 \equiv 0$, thus the rotational dynamics of the aerial vehicle and the one of the link are completely unrelated. The trajectory of $\elevation$ evolves only driven by the gravity force and the initial conditions of $\elevation,\delevation$, while $\pitch$ evolves depending  only on the initial conditions of $\pitch,\dpitch$ and the input torque $u_2$.

Notice that in this case, considering opportune higher order of differentiation we were able to prove the flatness with respect to $\by^b$ even for the singular case, i.e., $u_1 = 0$.
On the other hand, the singularity of the flat output $\by^a$ (i.e., $y_1^a + y_2^a = \pi/2 + k\pi$) it is  structural thus  impossible to eliminate with a similar technique.

\subsubsection{Discussion on the Sign of the Thrust}

From~\eqref{eqn:model:balanceForces_r} the following solution is also feasible
\begin{align*}
u_1 = 	-\norm{\mathbf{r}}, \qquad 
x_3 = \textrm{atan2}\left(-r_1/\|\mathbf{r}\|,-r_3/{\|\mathbf{r}\|}\right),
\end{align*}
which is the `opposite' of~\eqref{eqm:flatness:posSol}.
Therefore, from the same output  $\output^b$, one can get two possible different attitude and thrust pairs, $\bz_B = \pm \frac{\mathbf{r}}{\norm{\mathbf{r}}}$ and $u_1 = \pm \norm{\mathbf{r}}$. These solutions represent the same force vector $ \bthrust=-u_1 \bz_B$, obtained with the vehicle `facing up' and `facing down', respectively. 
In order to discriminate this ambiguity, exploiting the system continuity, the correct attitude and thrust sign are chosen selecting the closest configuration to the previous state.
Moreover, as long as the trajectory does not imply $\norm{\mathbf{r}} = 0$ for any $t$, the sign of $\thrust$ is constant. 

Even though propellers that can provide negative thrust are not of common use, some solutions exist such as, e.g.,   variable-pitch and symmetric propellers, with a constant and a time-varying speed of rotation, respectively. %
In the common case of aerial vehicles with only  positive thrust, one should provide desired trajectories that are either sufficiently distant from the condition $ \norm{\mathbf{r}} = 0$ (to have also some robustness against disturbances), or should plan a trajectory that avoids the change of the thrust sign when passing trough the condition $ \norm{\mathbf{r}} = 0$ , e.g., with a cusp or a non-zero time derivative for $\mathbf{r}$.

\section{State-Feedback Control}\label{sec:control}
A similar concept to differential flatness is \textit{exact feedback linearizability}~\cite{1995-FliLevMarRou}.
Differential flatness and exact dynamic feedback linearizability are correlated, indeed they are equivalent on an open and dense set of the state space~\cite{2002-DelOri}.
In this section we show that both $\output^a$ and $\output^b$ are also linearizing output for the system \eqref{fig:model:system_model}, and using the \textit{static} and \textit{dynamic feedback linearization} control techniques we can obtain input-output decoupling and full state linearization via static and dynamic feedback respectively. 
In the following two subsections we propose the design of the controllers for $\output^a$ and $\output^b$ respectively, whose global characteristics are summarized in \tab\ref{tab:control}.

\begin{table*}[t]
  \centering
  \resizebox{.99\textwidth}{!}{
  \begin{tabular}{ccrcccccc}
  \toprule
  \multirow{2}{*}{Controller}	&
  Controlled	&
  \multirow{2}{*}{Feedback Linearization}	&
  \multirow{2}{*}{Singularity}	&
  Tot. Relative	& 
  Internal		&
  Desired		&
  Uncontrolled  	&	
  Used		\\

  &
  Output &
  &
  &
  Degree&
  Dynamics		&
  Trajectories	&
  Variable		&
  Variables		\\
  \midrule
  $\Gamma^a$	&
  $\elevation,\pitch$ &
  Static: $\inp = [\thrust\;\, \torque]^T$ 	&
  $\elevation + \pitch = \frac{\pi}{2} + k\pi$ &
  $2 + 2 = 4$ &
  No	&
  $\elevation^d(t)\in C^1,\; \pitch^d(t)\in C^1$	&
  $\stress$ &
  $\state$	\\

  ${\Gamma^a}'$	&
  $\elevation,\pitch$ &
  Dynamic: $\inp' = [\dthrust\;\, \torque]^T$ 	&
  $\elevation + \pitch = \frac{\pi}{2} + k\pi$ &
  $3 + 2 = 5$ &
  No	&
  $\elevation^d(t)\in C^2,\; \pitch^d(t)\in C^1$	&
  $\stress$ &
  $\state$	\\

  $\Gamma^b$	&
  $\elevation,\stress$ &
  Dynamic: $\bar \inp = [\ddot\thrust\;\, \torque]^T$ 	&
  $\thrust = 0$ &
  $4 + 2 = 6$ &
  No	&
  $\elevation^d(t)\in C^3,\; \stress^d(t)\in C^1$	&
  $\pitch$	&
  $\state$
  \\
  \bottomrule  
  \end{tabular}
  }
  \caption{Controllers characteristics (without the observer).}
  \label{tab:control}
\end{table*}

\subsection{Controller for $\output^a$}\label{subsec:controller:a}

Considering the equation~\eqref{eqn:model:output:a} and applying the feedback linearization technique, we see that we have to derivate twice the output in order to see the input appear, obtaining the second and fourth rows of~\eqref{eqn:model}:
\begin{align*}
\ddot \output^a = \matrice{\modCon_1\cosS{x_1} \\ 0} + 
\matrice{\modCon_2\cosS{x_1 + x_3} & 0 \\ 0 & \modCon_3}\inp = \mathbf{b}^a(\state) + \mathbf{E}^a(\state)\inp. 
\end{align*}
Notice that the decoupling matrix $\decouplingM^a(\state)$ is invertible if and only if $x_1 + x_3 \neq \pi/2 + k\pi$ with $k\in\mathbb{Z}$. If this condition holds, assuming $\state$ known, the control law 
\begin{align}
\inp = \Gamma^a(\state,\virtualInput^a) = {\decouplingM ^a}^{-1}({\state}) \left[ -\mathbf{b}^a({\state}) + \virtualInput^a \right],	\label{eqn:control:a:input}
\end{align}
where $\virtualInput^a=[v^a_1\;\,v^a_2]^T$ are virtual inputs, brings the system in the form of a decoupled linear system $\matrice{ {y_1^a}^{(2)} & {y_2^a}^{(2)} }^T = 
\matrice{v^a_1 & v^a_2}^T = \virtualInput^a$.
Then, given the desired trajectories ${y_1^{a}}^d(t)\in C^1$ and ${y_2^{a}}^{d}(t)\in C^1$ for the link elevation $y^a_1$ and the vehicle attitude $y^a_2$ respectively, one can achieve the tracking designing a standard state-feedback linear controller.
Moreover the total relative degree\footnote{The sum, for each entry of the output, of the number of times that each output entry has to be differentiated in order to see at least one input appear.} 
$r^a = r_1^a + r_2^a = 2 + 2 = 4$ is equal to the dimension of the  state, thus there is no internal dynamics~\cite{1995-Isi}.

Notice that the control actions $\inp$ required by this controller may be discontinuous due to, e.g. desired trajectories possessing a discontinuity in the second derivative (which, in fact, are not forbidden by the $C^1$ requirement) or simply, in the real case, due to some noise in the measurements. This has to be taken into account in case one would like, e.g., to minimize mechanical vibrations of the link.
Furthermore, it has to be kept in mind that discontinuous inputs cannot be performed by the physical system in exam
because, e.g., the thrust produced by a propeller is an algebraic function of its angular speed and the angular acceleration (and the corresponding air flow change) cannot be infinite.
However, if one needs to enforce smoother inputs it possible to apply a dynamic compensator in order to get a sufficiently smooth control signal.

For example, if one needs at least a continuous thrust input $u_1$, we can assume as new control input the vector $\inp' = [\dot{u}_1\;\, u_2]^T$, considering the velocity of the thrust intensity as new controllable input, $\dot u_1 = \dot \thrust$. The output $y_1^a$ has to be differentiated three times to see the input appear
\begin{align*} 
\matrice{{y^a_1}^{(3)} \\ {y^a_1}^{(2)}} = \underbrace{\matrice{-\modCon_1\sinS{x_1}x_2 - \modCon_2\sinS{x_1 + x_3}(x_2 + x_4)u_1 \\ 0}}_{\textstyle\bar{\mathbf{b}}^a(\state,u_1)} + \mathbf{E}^a(\state)\inp'. 
\end{align*}    
The decoupling matrix does not change, thus, if $x_1 + x_3 \neq \pi/2 + k\pi$ with $k\in\mathbb{N}$, we can apply the control law
\begin{align}
 \inp' = {\Gamma^a}'(\state,u_1) = {\decouplingM^b}^{-1}(\state) \left[ -\bar{\mathbf{b}}^a(\state, u_1) + \bar\virtualInput^a \right],
\label{eqn:control:a:input:dynamic} 
\end{align}
that brings the system in the form of a decoupled linear system $\matrice{{y_1^a}^{(3)} & {y_2^a}^{(2)}}^T = 
	\matrice{v^a_1 & v^a_2}^T = \bar\virtualInput^a$.
Moreover the relative degree is still equal to the dimension of the extended state that includes also $u_1$, i.e., $\bar{r}^a = r^a + 1 = 5 = n + 1 $, thus the linearization is exact and without internal dynamics. 
This smoother form of the controller will be useful in order to close the control loop with the observer of \sect\ref{sec:observer}.

\subsection{Controller for $\output^b$}\label{subsec:controller:b}

Differently from the previous case, the static feedback with respect to the output $\output^b$ is not enough in order to exactly linearize the system. Indeed, the  decoupling matrix in this case results always singular because the input torque does not appear on both elevation's acceleration and link force, as instead the thrust intensity does. A dynamic compensator is then needed in order to delay the appearance of the thrust in the  output derivatives. 

As done in~\cite{2015a-TogFra} we consider as new input $\bar \inp = \left[ \ddot{u}_1\;\, u_2\right]^T=\left[ \bar u_1\;\, \bar u_2\right]^T$, considering the acceleration of the thrust intensity as new controllable input, $\ddot{u}_1 = \ddot{\thrust}$. Due to the dynamic compensator the actual state of the dynamical system is given by the extended vector $\bar{\state} = \left[ \elevation\;\, \delevation\;\, \pitch\;\, \dpitch\;\, u_1\;\, \dot{u}_1 \right]^T$, that contains also the thrust intensity and its derivative.\footnote{This does \emph{not} mean or imply that one has to measure or estimate $u_1$ or  $\dot{u}_1$ since they are easily and safely computed as integrals of the new input $\ddot{u}_1$.}    
In this condition the outputs $y_1$ and $y_2$ have to be differentiated respectively four and two times to see the new input appear
\begin{align}
	\matrice{y_1^{(4)} \\ y_2^{(2)}} = \mathbf{b}^b(\bar{\state}) + \underbrace{\matrice{\modCon_2 \cosS{x_1 + x_3} & -\modCon_2 \modCon_3 \sinS{x_1 + x_3}u_1 \\	\sinS{x_1 + x_3} & \modCon_3\cosS{x_1 + x_3}u_1}}_{ \textstyle \decouplingM^b(\bar{\state})}\bar \inp, \label{eqn:control:output:2thDerivative}
\end{align}
where $\mathbf{b}^b(\bar{\state})$ depends only to the extended state, and the decoupling matrix $\decouplingM^b (\bar{\state})$ is always invertible except for the case $u_1 \neq 0$. Assuming $u_1 \neq 0$ and $\bar\state$ known, the control law
\begin{align}
 \bar\inp = \Gamma^b(\bar{\state}) = {\decouplingM^b}^{-1}(\bar{\state}) \left[ -\mathbf{b}^b(\bar{\state}) + \virtualInput^b \right],	\label{eqn:control:b:input}
\end{align}
where $\virtualInput^b=[v^b_1\;\,v^b_2]^T$ are virtual inputs, brings the system in the form of a decoupled linear system $\matrice{{y_1^b}^{(4)} & {y_2^b}^{(2)}}^T = 
	\matrice{v^b_1 & v^b_2}^T = \virtualInput^b$.
Then, given a desired trajectory ${y_1^b}^{d}(t)\in C^3$ for the link elevation ${y_1^b}$ and a desired trajectory ${y_2^b}^{d}(t)\in C^1$ for the link force ${y_2^b}$, one can achieve the tracking by resorting to a standard state-feedback linear controller for $\virtualInput^b$.
Notice that the total relative degree $r^b = r^b_1 + r^b_2 = 4 + 2 = 6$ is equal to the dimension of the extended state, thus there is no internal dynamics also in this case. 
This ensures the stability of the attitude of the vehicle while following any desired trajectory $\output^{b\,d}(t)$ .

\subsection{Discussion on the Controllers}\label{subsec:controllers:ab}

The two control laws~\eqref{eqn:control:a:input}, and~\eqref{eqn:control:b:input} present a singularity in $x_1 + x_3 = \pi/2 + k\pi$ and $u_1 = 0$, respectively. 
For the controller $\Gamma^a$, the condition $x_1 + x_3 = \pi/2 + k\pi$, verified when the link and the thrust direction are parallel, corresponds to a series of parallel lines in the space of $x_1$ and $x_3$.
In practice this means that for a pair of given initial and final desired conditions, a trajectory that avoids the singularity might not exist (see also discussion in~Sec.~\ref{subsec:flatness:a}).
On the other hand, the singularity of the controller $\Gamma^b$ ($u_1 = 0$), corresponds to isolated points in the state space, thus easily avoidable with a planner.

If a desired trajectory passes through the singularity then the inversion of the decoupling matrix is not feasible and the controller cannot be applied. 
Moreover this represents an inconvenience also in the neighborhood of the singularity where small virtual inputs may cause large inputs. 
Nevertheless, if the singularity occurs for only an instant then the controller is still applicable in practice, e.g., approximating the control law using the \textit{damped least-square (DLS) inversion}~\cite{2009-SicSciVilOri}. i.e., in the neighborhood of the singularity, $\decouplingM^{-1}$ is approximated with $\decouplingM^\star = \decouplingM^T(\decouplingM\decouplingM^T + c^2\mathbf{I})^{-1}$.
It is proven that $\decouplingM^\star$ is the optimal solution in order to minimize the inversion error while bounding the norm of the input, where $c\in\nR{}$ is chosen to balance the two objectives. 
However, one has to plan trajectories that do not require to stay constantly in singularity. 

Observe that, in order to compute the control actions in~(\ref{eqn:control:a:input}), and~(\ref{eqn:control:b:input}), only the knowledge of the state is needed. 
Also the controller $\Gamma^b$ needs only the knowledge of the original state $\state$. Indeed $u_1,\, \dot u_1$ are known because are internal variables of the controller, and the outputs and their derivatives can be computed from the state as done, e.g., in~\eqref{eqn:model:output:b} and in the intermediate derivation steps that bring to~\eqref{eqn:control:output:2thDerivative}. 

Finally, these controllers satisfy only one of our requirements (decoupled tracking of the output) but they still need the full state information to be implemented, instead of just relying on the accelerometer and gyroscope measurements, as we want to be the case. This limitation is overcome in Sec.~\ref{sec:observer}.

\section{Observer and Measurement-Feedback Control}\label{sec:observer}

Measuring the whole state $\state$ using many sensors in order to compute the control action is often practically unfeasible due to, e.g., the costs and payload limitations of aerial robots. 
Furthermore, possible sensor failures call for the ability to still control the platform with a forcedly limited number of sensors. Driven by these two practical reasons and also by the intrinsic theoretical appeal of solving control problems with minimal sensing, in this section we show that the standard on-board inertial  sensors (i.e., an accelerometer plus a gyroscope) are sufficient to estimate the full state $\state$ of the system, and we show the design of an exact nonlinear observer for that purpose. Furthermore, we show that the combination of the observer and the controller results in an almost-globally asymptotically stable closed loop system.

The on-board inertial sensor equipment is composed by a gyroscope, that measures the angular velocity, and an accelerometer that measures the specific acceleration of the aerial vehicle, i.e., the linear acceleration of the CoM, minus the gravity, expressed in the body frame, i.e.,
\begin{align}
\omega &= x_4 \label{eqn:observer:gyro}\\
\accMeasures &= \mathbf{R}_W^B(\ddposR + g\bz_W) = \left[ a_x \;\, a_y \;\, a_z \right]^T, \label{eqn:observer:acc}
\end{align} 
where $\mathbf{R}_W^B = {\mathbf{R}_B^W}^T$. 
In the following we omit the second line which is  zero by construction, i.e., we assume $\accMeasures = \left[ a_x \;\, a_z \right]^T$.

Observing the full state of system~\eqref{eqn:model} using the partial measurements~\eqref{eqn:observer:gyro}, and~\eqref{eqn:observer:acc} is a nontrivial nonlinear observation problem, where the nonlinearities appear both in the system dynamics and in the measurements. We show in the following how this problem can be successfully tackled.

Fist of all exploiting~\eqref{eqn:observer:gyro} we can define the gyroscope measurement as a new input $u_3 = \omega$ that let us reduce the system dimension from four to three.
Then, substituting~\eqref{eqn:model:balanceForces} and~\eqref{eqn:model} into~\eqref{eqn:observer:acc} and performing some simple algebraic manipulation on the accelerometer measurement we obtain
\begin{align}
\matrice{\dot{x}_1 \\ \dot{x}_2 \\ \dot{x}_3} &= \matrice{ 0 & 1 & 0 \\
										0 & 0 & 0\\
										0 & 0 & 0 }\matrice{x_1 \\ x_2 \\ x_3} + %
		\matrice{ 0 \\ a_1 \cosS{x_1} + a_2 \cosS{x_1 + x_3}u_1 \\ u_3 }\label{eqn:observer:reducedModelDyn}\\
\accMeasures &= \matrice{ l\cosS{x_1 + x_3} \left( x_2^2 + \modCon_1\sinS{x_1} + \modCon_2\sinS{x_1 + x_3} u_1 \right)	 \\
		  l\sinS{x_1 + x_3}  \left( x_2^2 + \modCon_1\sinS{x_1} + \modCon_2\sinS{x_1 + x_3} u_1 \right)	 - \modCon_2 u_1}. \label{eqn:observer:reducedModelMeas}			
\end{align}
The problem is then `reduced' to the observation of the state $[x_1\;x_2\;x_3]^T$ from the knowledge of the measurements $\accMeasures$ and the control inputs $[u_1 \; u_3]^T$. 
However this problem is still nonlinear both in the system dynamics and in the measurement map.

In order to solve nonlinear observation problems there are mainly two classes of methods: approximate nonlinear observers and exact nonlinear observers. The first class relies on approximating the nonlinearities with linear or almost-linear maps around the current estimate, %
the main disadvantage being the local approximative nature of the methods. 
The second class of methods consists in nonlinear systems whose state is analytically  proven to converge to the real state of the original system. Designing such observers is in general much more difficult since  it is often hard to prove the asymptotic stability of a nonlinear  system.  However the observers of this class may guarantee (almost) global convergence, this is why we decided to search for an observer in the second class.

In the literature of exact nonlinear observers an important role is played by a particular class of systems known as \emph{in triangular from}, that, in the case of a three-dimensional state system like~(\ref{eqn:observer:reducedModelDyn}--\ref{eqn:observer:reducedModelMeas}) are written as:
\begin{align}
\begin{split}
\dot{\underline\state} & = \matrice{0 & 1 &  0 \\ 0 & 0 & 1 \\ 0 & 0 & 0}\underline\state + \matrice{0\\ 0\\ 1}\phi(\underline\state,\underline{\mathbf{u}}) + \boldsymbol{\lambda}(\underline{\mathbf{u}}) \\
\underline{w} &= \matrice{1 & 0  & 0	}\underline\state,
\end{split} \label{eqn:observer:desiredForm}
\end{align}
where $\underline\state\in\mathbb{R}^3$ is the state vector, $\underline\bu\in\mathbb{R}^{m_{\underline\bu}}$ is control input (with $m_{\underline\bu}\geq0$), $\underline{w}\in\mathbb{R}$ is the measurement and $\phi: \mathbb{R}^3\times\mathbb{R}^{m_{\underline\bu}}\to\mathbb{R} ,\; \boldsymbol{\lambda}: \mathbb{R}^{m_{\underline\bu}}\to\mathbb{R}^3$ are any nonlinear maps. For this class of nonlinear systems, in order to estimate the state one can use the nonlinear \textit{high gain observer} (HGO)\cite{2002-Kha} that ensures almost global convergence of the estimated state to the real one. 

Although at first view system~(\ref{eqn:observer:reducedModelDyn}--\ref{eqn:observer:reducedModelMeas}) does not resemble  to a system in triangular form like~\eqref{eqn:observer:desiredForm}, we shall demonstrate in the following that it can be put in that form using a few appropriate nonlinear transformations.

\subsection{State/Output Transformations and HGO Design}
\label{subsect:observer:HGO}

In the following we prove that there exist a change of coordinates from the original state  $\state$ to a new state $\bz=[z_1\;z_2\;z_3]^T$ and from the original measurements $\accMeasures$ to a new measurement $w$ such that the system~(\ref{eqn:observer:reducedModelDyn}--\ref{eqn:observer:reducedModelMeas}) appears in triangular form.
While doing so we highlight also the intuitions that led us to discover this particular  change of coordinates.

First of all since the term $x_1 + x_3$ occurs frequently in~\mbox{(\ref{eqn:observer:reducedModelDyn}--\ref{eqn:observer:reducedModelMeas})} a  simplifying choice is to assume $z_1 = x_1+x_3$.
With this choice we have $\dot z_1 = \dot x_1 + \dot x_3 = x_2 + u_3$ and therefore it is natural to choose $z_2 = x_2$ to obtain $\dot z_1 = z_2 + u_3$, thus matching with the first row of the sought triangular form~\eqref{eqn:observer:desiredForm}.

Now, if we compare the second and third rows of~\eqref{eqn:observer:reducedModelDyn}, with the corresponding rows of the sought triangular form~\eqref{eqn:observer:desiredForm}  we see that: 
\begin{inparaenum}[\em i)]
\item in the triangular form~\eqref{eqn:observer:desiredForm} the state-dependent nonlinearity $\phi$ appears only in the last row of the dynamics, but, on the other hand
\item in~\eqref{eqn:observer:reducedModelDyn} the nonlinearity appears already in the second row.
\end{inparaenum} Therefore, in order to push this `undesired' nonlinearity down from the second to the third row we can define $z_3 = \dot z_2 = \dot x_2 = a_1 \cosS{x_1} + a_2 \cosS{x_1 + x_3}u_1$.

Summarizing, we propose the following change of variables  
\begin{align}
\begin{split}
z_1 = x_1+x_3, \quad
z_2 = x_2,	\quad
z_3 = \dot x_2 = a_1 \cosS{x_1} + a_2 \cosS{x_1 + x_3}u_1, 
\end{split}\label{eqn:observer:stateTransf}
\end{align}
that transforms the system~\mbox{(\ref{eqn:observer:reducedModelDyn}--\ref{eqn:observer:reducedModelMeas})} in the following form
\begin{align}
\dot{\bz} &= \underbrace{\matrice{0&1&0 \\ 0&0&1 \\ 0&0&0}}_{\textstyle \mathbf{A}} \bz + \underbrace{\matrice{0 \\ 0 \\ 1}}_{\textstyle\mathbf{B}}\sigma'(\bz,\sinS{x_1},u_1,\dot u_1,u_3)  + \underbrace{\matrice{{u_3} \\ 0 \\ 0}}_{\textstyle \boldsymbol{\lambda}(u_3)}
	\label{eqn:observer:newState:model}	\\
	\accMeasures &= \matrice{ l\cosS{z_1} \left( z_2^2 + \modCon_1\sinS{x_1} + \modCon_2\sinS{z_1} u_1 \right)	\\
		  l\sinS{z_1}  \left( z_2^2 + \modCon_1\sinS{x_1} + \modCon_2\sinS{z_1} u_1 \right)	 - \modCon_2 u_1}, \label{eqn:observer:newState:acc}	
\end{align}
where the sole state-dependent nonlinearity $\sigma' = a_1 z_2 \sinS{x_1} + a_2\cosS{z_1}\dot u_1 - a_2\sinS{z_1}(z_2 + u_3)u_1$ is now appearing in the third row, as desired. 
Notice that we have, on purpose, left the term $\sinS{x_1}$ untransformed. %
In the following we show why this choice is convenient instead of directly computing $\sinS{x_1}$ from~\eqref{eqn:observer:stateTransf} .

To reach the form of~\eqref{eqn:observer:desiredForm} it remains to extract a measurement of $z_1$ from the accelerometer reading. From~\eqref{eqn:observer:newState:acc}, defining
\begin{align}
\begin{split}
\eta = \sqrt{a_x^2 + \left( a_z + \modCon_2 u_1  \right)^2} = \pm l \left( z_2^2 + \modCon_1\sinS{x_1} + \modCon_2\sinS{z_1} u_1 \right), \label{eqn:observer:eta}
\end{split}
\end{align} 
we can obtain a direct measure of $z_1$ writing
\begin{align}
\measure = \textrm{atan2}\left( \pm{a_x}/{\eta},\; \pm{(a_z + \modCon_2 u_1)}/{\eta} \right) = z_1 + k\pi	\label{eqn:observer:w},
\end{align}
where $k\in\{0,1 \}$.
Notice that the transformation is possible only if $\eta\neq 0$. 
In particular, from equations~\eqref{eqn:model:output:b} and~\eqref{eqn:observer:eta} it results that $\eta = \pm\frac{1}{\massR}\left( \frac{1}{\modCon_2}z_2^2 + \frac{\modCon_1}{\modCon_2}\sinS{x_1} + \sinS{z_1} u_1 \right) = \pm\frac{\stress}{\massR}$, thus the transformation~\eqref{eqn:observer:w} requires non zero force along the link, $\stress \neq 0$.
This correspondence highlights, intuitively, that the condition $\eta\neq 0$ it is not just  related to our particular transformation choice but is a structural observability requirement.
Indeed, if the force along the link is zero, the aerial vehicle and the link become two independent systems\footnote{In fact, due to the assumption that the mass and rotational inertia of the link are negligible, the link force represents the only coupling force between the two sub-systems.} and the onboard inertial sensor is not enough to estimate the entire state.
In the design of the observer we need to consider this singularity especially for the cases where the desired link force passes from a tension ($\stress > 0$) to a compression ($\stress < 0$).

The accelerometer is also used to replace $\sinS{x_1}$ into~\eqref{eqn:observer:newState:model} obtaining the sought triangular form~\eqref{eqn:observer:desiredForm}. In particular from~\eqref{eqn:observer:eta} we can write 
\begin{align}
\sinS{x_1} =\left( \pm{\eta}/{l} -z_2^2 - \modCon_2\sinS{z_1} u_1\right)/{\modCon_1}.	\label{eqn:observer:sinX1}
\end{align}
With reference to~\eqref{eqn:observer:newState:model} we can then write $\sigma' = \sigma(\bz,\bInObs)$, i.e., in terms of only $\bz$ and  known quantities $\bInObs = [u_1\; \dot{u}_1\; u_3\; \pm\eta]^T$.

Observe that, the sign to be put in front of $\eta$ is ambiguous. It is convenient to recast this ambiguity putting always a positive sign and considering two possible values:  $\eta_+ = +\eta$ and $\eta_-= -\eta$. 
Corresponding to these quantities we then get two different dynamic models, the one with $(\sigma(\bz,\bInObs_+),\measure_+)$ and the one with $(\sigma(\bz,\bInObs_-),\measure_-)$ corresponding to the use of $\eta_+$ and $\eta_-$, respectively.
At each time instant only one choice for $\eta$ is correct, i.e., $\eta = \stress / \massR$, to which corresponds the correct measure $\measure = z_1$ and the correct dynamics. 
It is not possible, however, to discriminate the correct choice instantaneously from the measurements only. 
In \sect\ref{subsec:observer:measurementDisambiguation} we propose a discriminating solution based on the prediction error.

Although less intuitive, the computation of $\sinS{x_1}$ into~\eqref{eqn:observer:newState:model} exploiting the accelerometer readings, allows to concentrate the ambiguity only on the sign of $\eta$ obtaining two possible dynamic models. 
Whereas with the more canonical technique, i.e., inverting the state transformation, we would have four different dynamic models. Indeed from~\eqref{eqn:observer:stateTransf} we would get
$\sinS{x_1} = \pm\sqrt{1-(z_3 - \modCon_2 \cosS{z_1} u_1)^2/\modCon_1^2 }$
that presents another ambiguity on the sign thus generating four possible combinations of the dynamics and measurement equations.

For the design of the observer let us assume that the correct choice between $\eta_+$ and $\eta_-$ is known.
In \sect\ref{subsec:observer:measurementDisambiguation} we shall propose a method to gain this knowledge.
Under this assumption, the described transformation of the state and the measurements have finally transformed the original system~\mbox{(\ref{eqn:observer:reducedModelDyn}--\ref{eqn:observer:reducedModelMeas})} in an equivalent one in a triangular form, i.e., $\dot{\bz} = \mathbf{A} \bz + \mathbf{B} \sigma(\bz,\bInObs)  + \boldsymbol{\lambda}(\bInObs)$ and $\measure = \matrice{1& 0& 0}\bz = \mathbf{C}\bz$, for which we can use the following high gain observer~\cite{2002-Kha}
\begin{align}
\dot{	\hat\bz} = \mathbf{A}\hat\bz +  \mathbf{B} \sigma(\hat\bz,\bInObs) + \boldsymbol{\lambda}(\bInObs) + \mathbf{H}({\measure} - \mathbf{C}\hat\bz),\label{eqn:observer:HGO}
\end{align}
where $\mathbf{H} = \matrice{\frac{\alpha_1}{\epsilon}\, \frac{\alpha_2}{\epsilon^2}\, \frac{\alpha_3}{\epsilon^3}}^T$, $\epsilon \in \nR{}_{>0}$, and $\alpha_i \in \nR{}_{>0}$ are set such that the roots of $p^3 + \alpha_1p^2 + \alpha_2p +\alpha_3$ have negative real part\footnote{The difference $(\measure - C\hat{\bzeta})$ stands here for the unique angle $\beta\in(-\pi,\pi]$ such that $\beta+C\hat{\bzeta}= \measure +k 2\pi$ for a certain  $k\in\mathbb{Z}$.}.

\subsection{Observation of Original State $\hat\state$}

\begin{figure}[t]
	\centering
	\includegraphics[width=0.87\columnwidth]{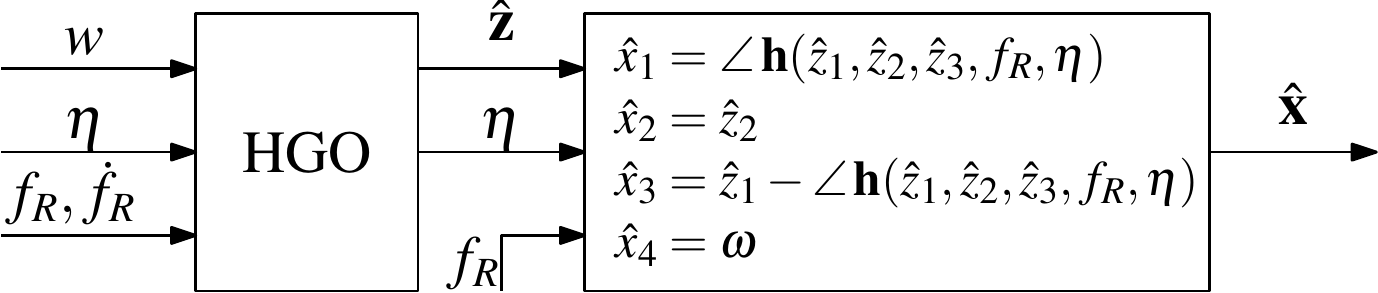}
	\caption{Observer.}
	\label{fig:observer}
\end{figure}

From the estimation of $\bz$, in order to obtain the estimation of the original state $\state$, we note that the state transformation \eqref{eqn:observer:stateTransf} is not directly invertible.
One can notice that the only knowledge of $\bz$ is not enough to retrieve $x_1$, indeed from~\eqref{eqn:observer:stateTransf} one can extract only $\cosS{x_1}$. 
Nevertheless, exploiting also the accelerometer measurements~\eqref{eqn:observer:sinX1}, we can write
\begin{align*}
\begin{bmatrix} \cosS{x_1}\\ 	\sinS{x_1} \end{bmatrix} 	=
	\begin{bmatrix} \frac{1}{{\modCon_1}} (z_3 - \modCon_2 \cosS{z_1} u_1)\\
	\frac{1}{\modCon_1}\left(\frac{\eta}{l} -z_2^2 - \modCon_2\sinS{z_1} u_1\right) 
	\end{bmatrix} =  \bh(\bz ,u_1,\eta).
\end{align*}
The state $x_1$ can then be computed as the phase of the unit vector $ \bh(\bz ,u_1,\eta)$ denoted by $\angle \, \bh(\bz,u_1,\eta)$. 
Thus the  estimation of the original state is given by
\begin{align}
\hat{\state} = \hat{\state}(\hat\bz,u_1,\eta) = \matrice{\angle \, \bh(\hat\bz,u_1,\eta)\\ 
\hat z_2 \\ 
\hat z_1 - \angle \, \bh(\hat\bz,u_1,\eta)\\ 
u_3 }. \label{eqn:observer:originalState} 
\end{align}
The full observer chain is then depicted in \fig\ref{fig:observer}.

\subsection{Closed-loop System Stability with State Observation}

We show now how the estimated state can be then used to compute the control action, closing the control loop with only on-board sensors, preserving the stability of the global system.

First of all notice that if controller $\Gamma^a$ in~\eqref{eqn:control:a:input} is used to control $\output^a$ then it is not possible to close the loop with the observer~\eqref{eqn:observer:HGO}. In fact, the observer requires the knowledge of $\dot{u}_1$, while the static controller $\Gamma^a$ directly computes the thrust $u_1$ without passing first through its derivative, and, even worse, $u_1$ can be even discontinuous and so non-differentiable. 
Nevertheless, one can use the controller ${\Gamma^a}'$ in~\eqref{eqn:control:a:input:dynamic}, thus obtaining a  differentiable $u_1$ and the knowledge of $\dot{u}_1$ that becomes an available internal state of the controller. Summarizing, ${\Gamma^a}'$ can be used with the observer while  ${\Gamma^a}$ can be used only with a full  knowledge of the state coming from the measurements. 

Let us now analyze the closed-loop system composed by the plant~\eqref{eqn:model} and the output feedback controller~\eqref{eqn:control:a:input:dynamic} computed on the basis of the estimated state $\hat\state$.
For~\eqref{eqn:observer:originalState} we can write
\begin{align}
 \inp' =  {\Gamma^a}'(\hat{\state}) = {\Gamma^a}'(\hat{\bz},u_1,\eta),\label{eqn:observer:controller:a}
\end{align} 
i.e., the control law depends directly on the estimation of the HGO, the measures and the state of the dynamic compensator. 
Since the controller is almost globally exponentially convergent, except the case $\elevation + \pitch = \pi/2 +k\pi$, then there exists $\epsilon^\star > 0$ such that, for every $0< \epsilon < \epsilon^\star$, the closed loop system with controller~\eqref{eqn:observer:controller:a} and observer~\eqref{eqn:observer:HGO} is exponentially convergent for every state except the cases in which either $\elevation + \pitch = \pi/2 +k\pi$ or $\stress=0$~\cite{2002-Kha}.

Similarly, for the output feedback controller~\eqref{eqn:control:b:input} computed using the estimated state $\hat\state$ there exists $\epsilon^\star > 0$ such that, for every $0< \epsilon < \epsilon^\star$, the closed loop system with controller
\begin{align*}
 \bar\inp =  \Gamma^b(\hat{\state},u_1,\dot u_1) = \bar{\Gamma}^b(\hat{\bz},u_1,\dot u_1,\eta)
\end{align*}
 and observer~\eqref{eqn:observer:HGO} is exponentially convergent for every state except the cases in which either $\thrust = 0$ or  $\stress = 0$. 

Although the convergence of the observer is almost global, an initialization phase of the estimation can be useful in order to minimize the transient duration, e.g., using the method proposed in~\cite{2013-LupDan} for a quasi static initial condition.

\subsection{Disambiguation of $\eta$ and Observability Discussion}	\label{subsec:observer:measurementDisambiguation}

\begin{figure}[t]
	\centering
	\includegraphics[width=0.9\columnwidth]{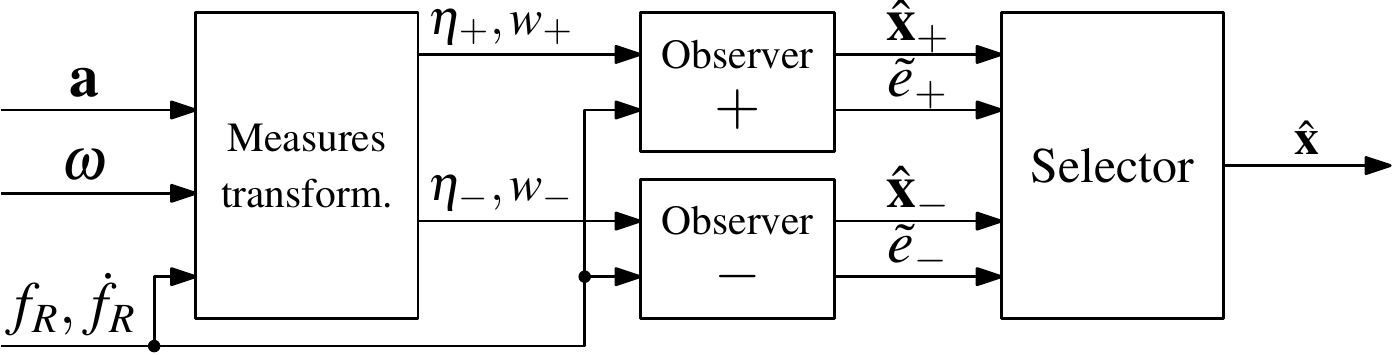}
	\caption{Global observer with disambiguation of $\eta$.}
	\label{fig:observerGlobal}
\end{figure}

As explained before, the transformation of the measurements presents an ambiguity on the sign of $\eta$, that can be considered positive, $\eta_+$, or negative, $\eta_-$. We show in this section how the correct choice can be easily made. Refer to~\fig\ref{fig:observerGlobal} for a graphical representation.

For each of the two possible choices let us implement an HGO equal to~\eqref{eqn:observer:HGO}
\begin{align*}
\dot{	\hat\bz}_+ &= \mathbf{A}\hat\bz_+ +  \mathbf{B} \sigma(\hat\bz_+,\bInObs_+) + \boldsymbol{\lambda}(\bInObs_+) + \mathbf{H}({\measure_+} - \mathbf{C}\hat\bz_+) \\
\dot{	\hat\bz}_- &= \mathbf{A}\hat\bz_- +  \mathbf{B} \sigma(\hat\bz_-,\bInObs_-) + \boldsymbol{\lambda}(\bInObs_-) + \mathbf{H}({\measure_-} - \mathbf{C}\hat\bz_-)
\end{align*}
obtaining two different estimations of the state $(\hat{\bz}_+,\hat{\bz}_-)$, and therefore two different estimations of the original state $(\hat{\state}_+,\hat{\state}_-)$ of which only one is correct.
In order to select the correct state we propose a discrimination method based on the comparison of the measurement prediction errors. 
At the first observer we assign a prediction error $\tilde e_+$ smoothed with an exponential discount factor: $\dot{\quality}_+ = \lambda \left( \norm{\accMeasures - \hat{\accMeasures}_+} - \quality_+\right)$,
where $\lambda \in \nR{}_{>0}$ sets the discount rate, and $\hat{\accMeasures}_+$  is defined as
\begin{align}
		\hat\accMeasures_+ = \matrice{ l\cosS{\hat x_{1+} + \hat x_{3+}} \left( \hat x_{2+}^2 + \modCon_{1}\sinS{\hat x_{1+}} + \modCon_{2}\sinS{\hat x_{1+} + \hat x_{3+}} u_{1} \right)	\\
		  l\sinS{\hat x_{1+} + \hat x_{3+}}  \left( \hat x_{2+}^2 + \modCon_{1}\sinS{\hat x_{1+}} + \modCon_2\sinS{\hat x_{1+} + \hat x_{3+}} u_{1} \right)	 - \modCon_2 u_1} \hspace{-0.2em}.
\label{eqn:observer:predictedMeasures}
\end{align}
The other prediction error $\quality_-$ is defined similarly.
Then we select the estimation provided by the observer implementation which produces the prediction error closer to zero, i.e., $\hat{\state} =\hat{\state}_+$ if $\quality_+ \leq \quality_-$ and $\hat{\state} =\hat{\state}_-$ otherwise.

In \fig\ref{fig:observerGlobal} the full observer with discrimination chain is represented.
Notice that the disambiguation of the two observers is not done directly using $\hat\bz$ because is not possible to write the predicted measure $\hat{\accMeasures}$ as function of $\hat{\bz}$ without introducing another ambiguity. 
Indeed, as we saw in \sect\ref{subsect:observer:HGO}, trying to replace $\hat\state$ with $\hat\bz$ into~\eqref{eqn:observer:predictedMeasures} inverting the state transformation~\eqref{eqn:observer:stateTransf}, we introduce an ambiguity on the sign of $\sinS{x_1}$. 
Whereas the problem does not hold if we apply this discrimination technique on the original state estimation $(\hat{\state}_+,\hat{\state}_-)$.

Notice that for the motions that implies a constant elevation ($x_2 = 0$), it is not possible to discriminate the correct observer. 
Indeed, with $x_2 = 0$ and since $\measure_{\pm} = z_1 + k\pi$, after a transient the two observers converge to ($\hat{z}_{1\pm} = z_1 + k\pi$,	$\hat{z}_{2\pm} = 0$, $\hat{z}_{3\pm} = 0$),
and, using~\eqref{eqn:observer:originalState}, we obtain ($\hat{x}_{1\pm} = x_1 + k\pi	$, $\hat{x}_{2\pm} = 0$, $\hat{x}_{3\pm} = x_3$, $\hat{x}_{4\pm} = x_4$).

Under this condition, from equation~\eqref{eqn:observer:predictedMeasures} it results that the prediction errors of the two observers converge both to zero thus making ineffective the  proposed discrimination strategy. 
Nevertheless, in practice this is not a problem.
Indeed, if the controller loop is closed with the wrong observer then the wrong estimation will let the control implement a law that is different from the sole that keeps $x_2 = 0$ causing $x_2 \neq 0$ and thus the predictions errors will become discriminant.

Finally, notice that the ambiguity issue discussed in this section is present only in the initial phases. Whenever the good observer is selected with sufficient certainty, one can switch off the other. 
For this purpose one can set a confidence threshold on the tracking error of the desired output. If an observer reaches the confidence threshold then this is identified as the correct one and the other one is switched off.

\subsection{Discussion on the Proposed Observer}
\subsubsection{Applicability}

If the link force $\stress$ is zero then $\eta$ is also zero and $\measure$ cannot be determined.
In the very special case that the link force has to be constantly equal to zero, the proposed observer is not applicable\footnote{The static observer in~\cite{2013-LupDan} requires a nonzero link force as well to work.} (while the controller would still work).
Nevertheless, if the desired link force is passing through zero for a sufficiently short time interval, then the proposed observer can be still used in practice by updating the filter without the correction term in that time instants, i.e., imposing
\begin{align*}
\dot{	\hat\bz} = \mathbf{A}\hat\bz +  \mathbf{B} \sigma(\hat\bz,\bInObs) + \boldsymbol{\lambda}(\bInObs) \quad \text{if } \measure=0.
\end{align*}
In this way the error dynamics becomes non strictly stable for a short moment but the dynamics returns asymptotically stable as soon as the link force returns to a non-zero value. 

\subsubsection{Robustness}

In order to deal with known drawbacks of the HGO, such as peaking phenomenon and noise sensitivity, many common practical solutions have been presented in the literature, see e.g.,~\cite{2002-Kha}. For example, to overcome the peaking phenomenon, it is sufficient to saturate the estimated state on a bounded region that defines the operative regions of the state for the system in exam. In the presence of measurement noise, the use a switched-gain approach can guarantee a quick convergence to the real state during the first phase while reducing the noise effects at steady state.

\section{Numerical Validation}\label{sec:simulations}

In this section we present a comprehensive  validation of the method with the main purpose of testing it under difficult non-ideal conditions, thus showing both its strengths and possible limits when applied on a real system.
We simulate a standard aerial vehicle whose nominal parameters are $\massR=1 [\rm kg]$, $\inertiaR = 0.25 [\rm kg\,m^2]$ and a link of length $l = 2 [\rm m]$.
Concerning the controller $\Gamma^b$ we set the gains of the linear outer control loop, $\mathbf{k}_1^b$ and $\mathbf{k}_2^b$, such that the error dynamics of $y_1^b$ and $y_2^b$ has poles in $(-1,-1.5,-2,-2.5)$ and $(-1,-1.5)$ respectively. While for the controller $\Gamma^{a'}$ we set the gains $\mathbf{k}_1^{a'}$ and $\mathbf{k}_2^{a'}$, such that the error dynamics of $y_1^a$ and $y_2^a$ has poles in $(-0.5,-1,-1.5)$ and $(-0.5,-1)$ respectively.
For the gains of the observer we set $\epsilon = 0.1$ and $(\alpha_1,\alpha_2,\alpha_3)$ such that the root of $s^3 + \alpha_1s^2 + \alpha_2s +\alpha_3$ are $(-6,-4.5,-3)$.
Those values guarantee the stability of the closed loop system and a sufficiently rapid convergence of the observer and controller.
For the controller $\Gamma^{a'}$, the desired trajectory is a smooth step, continuous up to the third order for $\elevation$ and up to the second order for $\pitch$, from the initial values $\elevation_0^d = 10^\circ$, $\pitch_0^d = 30^\circ$ to the final values $\elevation_f^d = 50^\circ$, $\pitch_f^d = 5^\circ$, respectively.
For the controller $\Gamma^{b}$, the desired trajectory is a smooth step, continuous up to the fourth order for $\elevation$ and up to the second order for $\stress$, from the initial values  $\elevation_0^d = 45^\circ$, ${\stress}_0^d = 3[\rm N]$, to the final values $\elevation_f^d = 135^\circ$, ${\stress}_f^d = 5[\rm N]$, respectively.  Smooth step-like trajectories (see Fig.~\ref{fig:sim:noise:b}), as it will be clear later, have the benefit of clearly showing the performances of the controllers  under three important conditions: the initial transient, the tracking of a fast time-varying signal, and the steady state. 

To obtain a complete validation, in the following we show a concise summary of the results about the stability and robustness of the proposed method under different non-ideal conditions, such as:
\begin{inparaenum}[a)]
\item nonzero initial tracking and estimation errors,
\item parametric variations,
\item generic CoM position and non-negligible link mass,
\item noisy sensor measurements, and
\item non-ideal motors.
\end{inparaenum} 

Although not very relevant for the investigation of the robustness of the method, we provide in \cite{2017-TogFra} extra plots relative to the performances of controller and observer for each non-ideal case with some typical values.
\subsection{Validation for Nonzero Initial Tracking/Estimation Errors}\label{subsec:generalModel}

In order to show the asymptotic convergence performances of both the controller and the observer we initialize the control system with nonzero initial tracking and estimation errors. 
After the convergence of the observer, which takes less than one second, the controller $\Gamma^{b}$ is able to steer the outputs along the desired trajectories with zero error.
A similar behavior is obtained for the controller $\Gamma^{a'}$.
Since the results only show that the controller is asymptotically convergent, as expected, we have reported the plots in~\cite{2017-TogFra} for limits of space.
\subsection{Parametric Variations}\label{subsec:paramVar}   
\begin{figure}[t]
	\centering
	\captionsetup[subfigure]{farskip=0.5em,
							 captionskip=-1em,
							 labelformat=empty,
							 width=\columnwidth}
	\subfloat[][]{\includegraphics[width=\columnwidth]{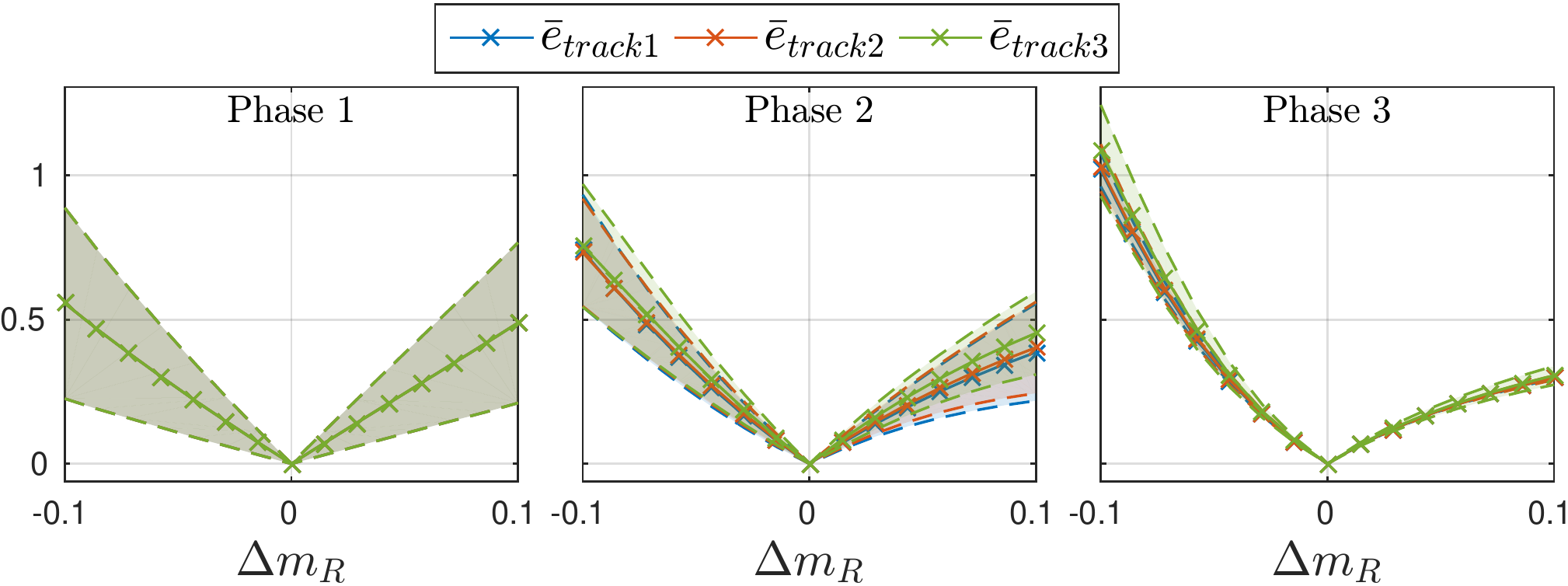}
	\label{fig:sim:paramVar:a:mR}}\\ 
	\subfloat[][]{\includegraphics[width=\columnwidth]{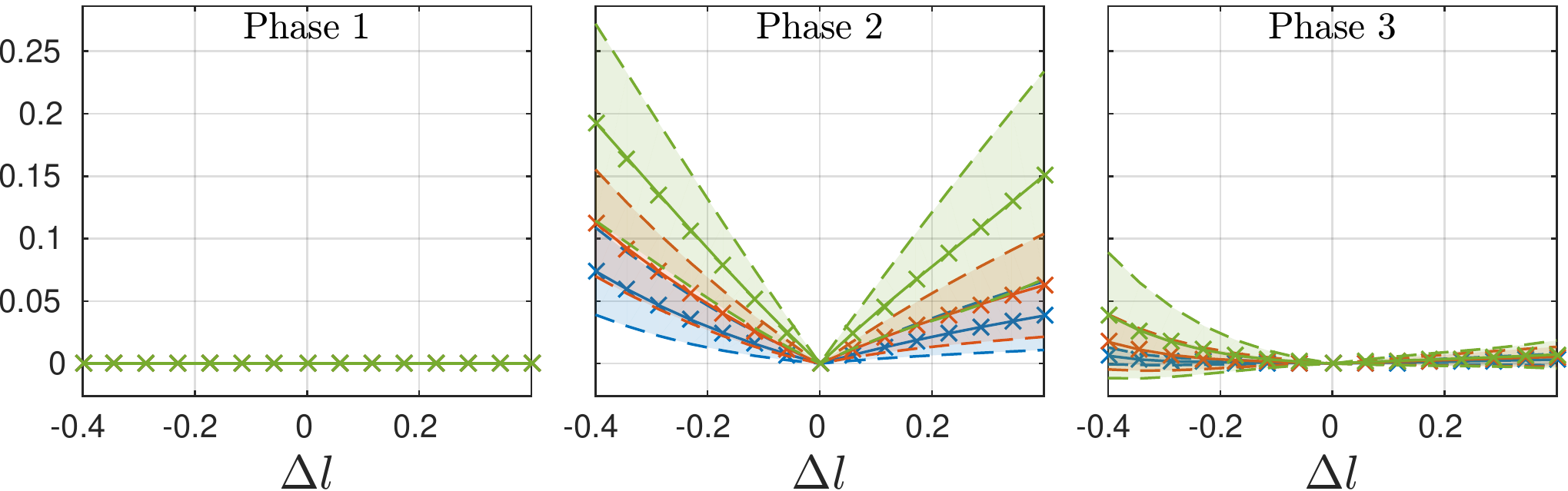}
	\label{fig:sim:paramVar:a:l}}\\	
	\subfloat[][]{\includegraphics[width=\columnwidth]{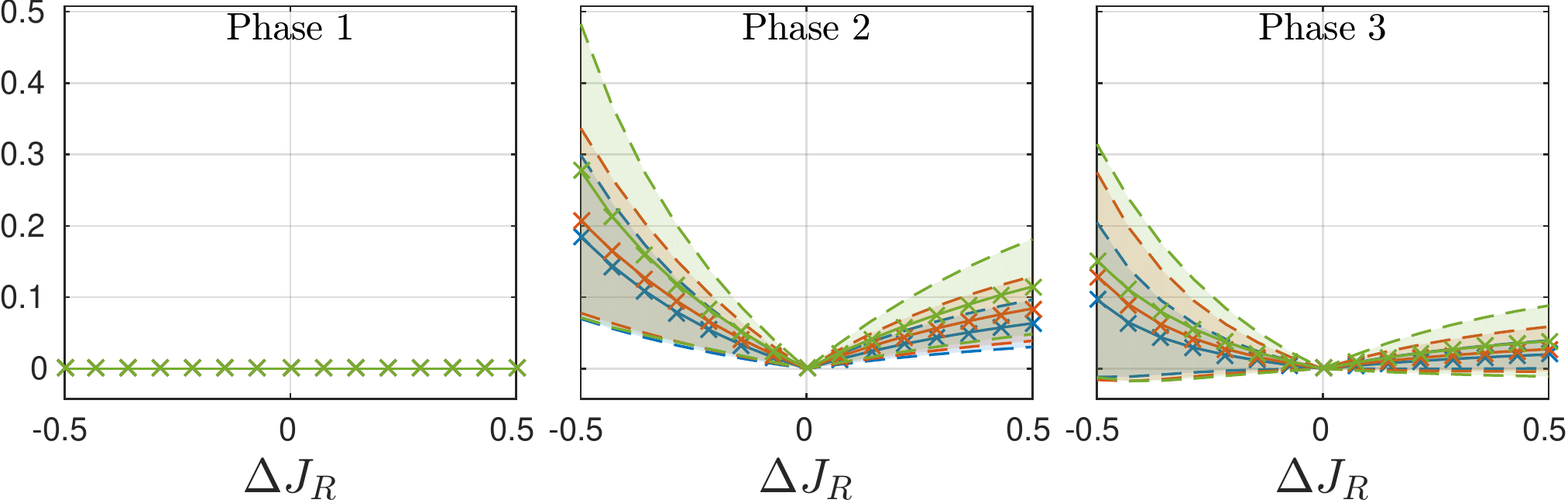}
	\label{fig:sim:paramVar:a:JR}}\\
	\caption{\emph{Parametric variation - Controller $\Gamma^{a'}$(elevation and attitude)}. The subscript 1, 2, and 3 correspond to the three different trajectory times. Outside of the displayed range of parametric variation the performances are unacceptable or the closed loop system results to be even unstable.}
	\label{fig:sim:paramVar:a}
\end{figure}
\begin{figure}[t]
	\centering
	\captionsetup[subfigure]{farskip=0.5em,
							 captionskip=-1em,
							 labelformat=empty,
							 width=\columnwidth}
	\subfloat[][]{\includegraphics[width=\columnwidth]{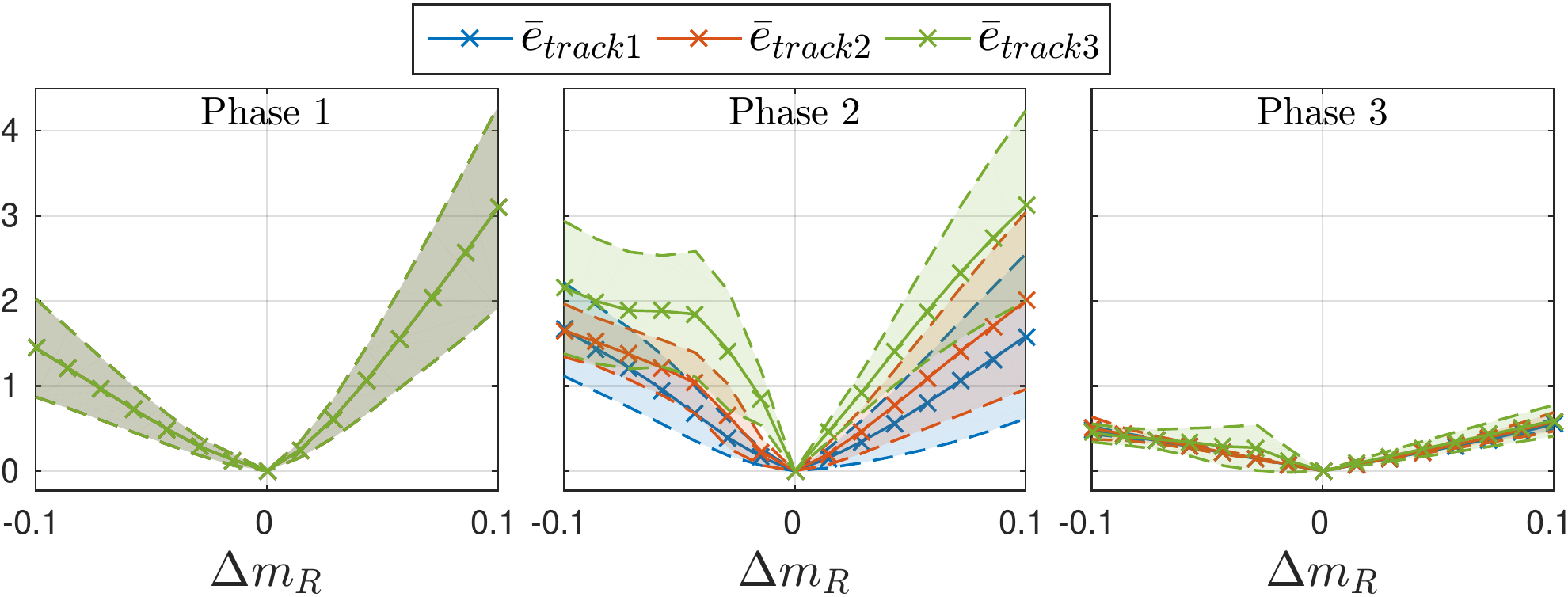}
	\label{fig:sim:paramVar:b:mR}}\\ 
	\subfloat[][]{\includegraphics[width=\columnwidth]{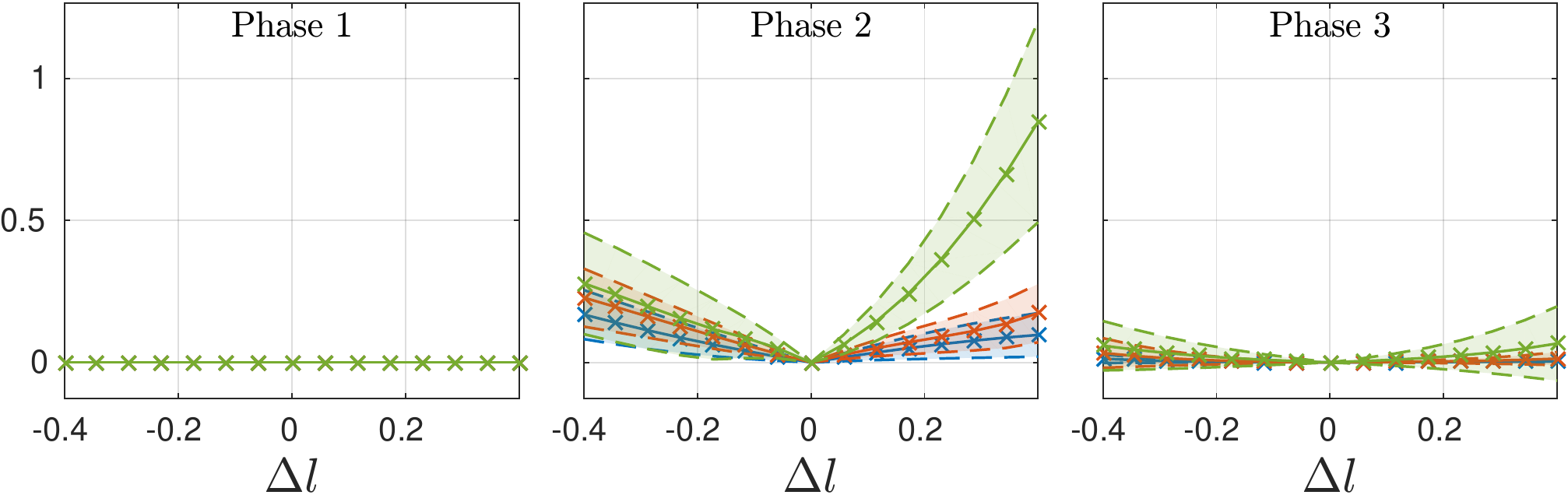}
	\label{fig:sim:paramVar:b:l}}\\	
	\subfloat[][]{\includegraphics[width=\columnwidth]{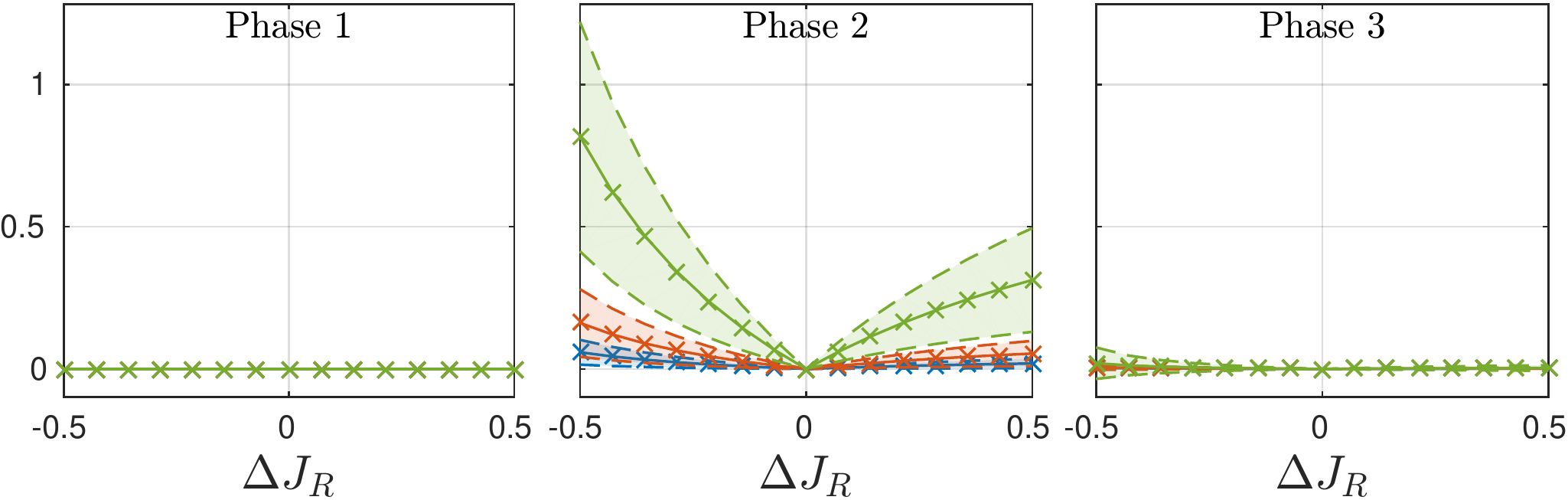}
	\label{fig:sim:paramVar:b:JR}}\\
	\caption{\emph{Parametric variation - Controller $\Gamma^{b}$ (elevation and link force)}.}
	\label{fig:sim:paramVar:b}
\end{figure}

In a real scenario the exact parameters of the system are unknown. The purpose of this section is to investigate the robustness of the proposed method in this scenario.
In particular we introduce discrepancies between the real model and the nominal one used in the controller/observer.  

We notice that in principle one could try to design an adaptive control law that is able to compensate for parametric uncertainties. However, this is clearly a tough objective, because the system is nonlinear and the available measurements are only the (nonlinear) accelerometer and the gyroscope readings. Therefore this goal is left as future work. Instead, we concentrate in this section on assessing the ranges of parameter variations that causes a degradation of the performance that remains within an acceptable bound. By doing so, we shall see in fact that the proposed control scheme possesses a remarkable robustness even without the presence of an adaptive design. 

Considering $l_0$,${\massR}_0$ and ${\inertiaR}_0$ the real parameter value and $l$,${\massR}$ and ${\inertiaR}$ the nominal one, we set  $l = (1 + \Delta l)l_0$, $(1 + \Delta \massR){\massR}_0$ and $(1 + \Delta \inertiaR){\inertiaR}_0$, where $\Delta\massR$, $\Delta l$ and $\Delta\inertiaR$ denote the relative parametric variations.

For obtaining a comprehensive analysis 
we tested the behavior for several different parametric variation combinations.
The results are plotted in Figs.~\ref{fig:sim:paramVar:a} and~\ref{fig:sim:paramVar:b}, where we show the mean tracking error, $\bar{e}_{track}$, and the corresponding standard deviation $\sigma_{\bar{e}_{track}}$,  defined as
\begin{align*}
e_{track}(t) &= \frac{\norm{y_1^d(t) - y_1(t)}}{y_1^d(t)} + \frac{\norm{y_2^d(t) - y_2(t)}}{y_2^d(t)} \\
\bar{e}_{track} &= \frac{1}{t_f-t_0}\int_{t_0}^{t_f} e_{track}(t) dt \\
\sigma_{\bar{e}_{track}} &= \sqrt{\frac{1}{t_f- t_0}\int_{t_0}^{t_f} (e_{track}(t) - \bar{e}_{track})^2 dt},
\end{align*}
where $t_0$ and $t_f$ are the initial and final time, respectively.
In the plots the solid line corresponds at the mean tracking error while the dashed lines correspond at the mean tracking error plus and minus its standard deviation.

The effect of an unknown parameter could also change with respect to the trajectory and in particular with respect to the velocity and acceleration at which the path is followed.
Consequently we plotted the mean tracking error, $\bar{e}_{track1}$, $\bar{e}_{track2}$ and $\bar{e}_{track3}$, for the same type of desired path (smooth step) but executed with  three different durations (increasing velocity): 	
\begin{inparaenum}	
\item $7 [\rm s]$,	
\item $5 [\rm s]$ and
\item $3 [\rm s]$ respectively.
\end{inparaenum}
We also analyze the error behavior dividing the trajectory into three phases: in the \textit{Phase 1}  (transient) the desired trajectory is constant and the analysis is more focused on the convergence of the observer; 
\textit{Phase 2} constitutes the dynamic part where the desired trajectory quickly goes from the initial value to the final one; 
the \textit{Phase 3}, the last, corresponds to the steady state condition where the desired trajectory is again constant.  
We show the tracking error for each of the three phases to better understand if a parameter variation affects more the transient, the dynamic phase, or the static one.

From \fig\ref{fig:sim:paramVar:a} and \fig\ref{fig:sim:paramVar:b} one can notice that, as expected, the performances get worse increasing the parametric variation. 
Furthermore, the same variation has more effect if the trajectory is more ``aggressive'' and it is followed with higher speed ($\bar{e}_{track3}$).
This is due to the fact that with higher speed and acceleration the inertial and Coriolis/centripetal terms become larger, and thus also the error in the feedback linearization increases, which in turn implies a worst tracking. %

Comparing the performances between the two controllers, we noticed that the controller  $\Gamma^{a'}$ results to be more robust then the controller $\Gamma^b$ in term of mean tracking error. 
This is due to the fact that for the controller $\Gamma^{a'}$, the dynamics of one of the controlled outputs, namely $\pitch$, is not influenced by the parameters such as mass and length of the link. 
This means that any variation on these parameters does not generates a worse tracking of $\pitch^d$, which results in a lower tracking error.

One can also notice that the mean tracking error is not in general symmetric with respect to the sign of the relative parametric variation.
For example for the controller $\Gamma^{a'}$ it is better to overestimate  the mass, and the length of the link rather than underestimating them, while for controller $\Gamma^b$ it results to be the opposite, even if these consideration are more relevant for the dynamic phase. 
Indeed, during the steady state phase the behavior is almost symmetrical.

Another fact that appears clear from the plots is that the variation that most influences the performances is the one on the mass of the aerial vehicle.
Fortunately, in practice this parameter can be easily measured with high precision. %

\subsection{General Model with Generic CoM Position and Non-negligible Link Mass}\label{subsec:generalModel}

\begin{figure}[t]
	\centering
	\includegraphics[height=0.37\columnwidth]{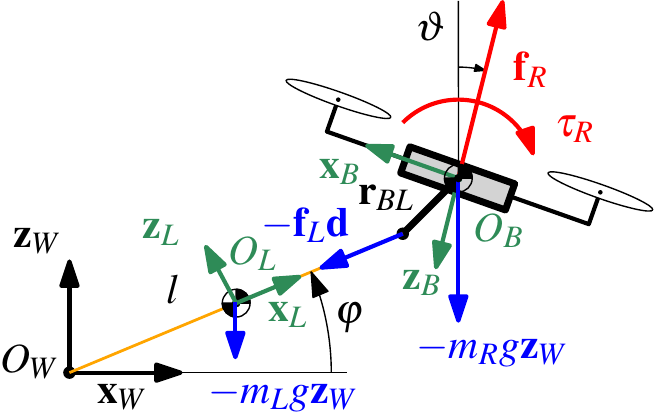}
	\caption{Representation of the more general system and its variables.}
	\label{fig:model:generalSystem}
\end{figure}

The controllers developed in this paper assume that the system can be represented with the model given in \fig\ref{fig:model:system_model}, where the CoM of the aerial vehicle coincides with the attachment point of the link to the vehicle and the link has a negligible mass. 
\fig\ref{fig:model:generalSystem} represents instead a more general model for which the assumptions done in \sect\ref{sec:model} are not fulfilled. 
Taking into account the definitions made in \sect\ref{sec:model} we then define a \textit{body frame}, $\lFrame$, attached to the link, with axes $\{ \bx_L, \by_L, \bz_L\}$ and origin $O_L$ coinciding with the center of mass (CoM) of the link. 
The position of $O_L$,  defined in $\wFrame$, is denoted with $\mathbf{p}_L = \left[x_L\;\, y_L\;\, z_L\right]^T $.
As for $\bFrame$ we have that $\by_L\equiv\by_B\equiv\by_W$ and $y_L\equiv0$.
For the validation we model the link as a rigid body of mass $\mass_L\in\nR{}_{>0}$ and inertia $\inertia_L\in\nR{}_{>0}$. 
Considering the inertia of the link as the inertia of an infinitesimally thin rigid tie with uniform distributed mass, we have also that $\inertia_L = \mass_L l^2 / 12$.
Assuming links with high stiffness, the deformations and the elongations results negligible with respect to the length of the cable itself, in the range of forces of our concern.
Therefore the link length is fixed. 
The link is connected  at one end to a fixed point coinciding with  $O_W$ and at the other end to a  point rigidly attached to the aerial vehicle whose  constant position in $\bFrame$ is denoted with $\offsetCoM = [r_x\,\; 0 \,\; r_z]^T$. 
If $\norm{\offsetCoM} = 0$ then the link is directly attached to the CoM of the aerial vehicle.

The mechanical model of the more general robotic system can be then derived writing the dynamics as the one in~\eqref{eqn:model:lagrangianModel} plus a disturbance due to the non idealities:
\begin{align*}
\mathbf{M}(\genCoordinates)\ddot{\genCoordinates} + \mathbf{g}(\genCoordinates) + \disturbance(\genCoordinates,\dot\genCoordinates,\ddot\genCoordinates,\inp) = \mathbf{Q}(\genCoordinates)\inp,
\end{align*}
where
\begin{gather*}
\disturbance(\genCoordinates,\dot\genCoordinates,\ddot\genCoordinates,\inp) = \bar{\mathbf{M}}(\genCoordinates)\ddot{\genCoordinates} + \bar{\mathbf{c}}(\genCoordinates,\dot{\genCoordinates}) + \bar{\mathbf{g}}(\genCoordinates) - \bar{\mathbf{Q}}(\genCoordinates)\inp,\\
\bar{\mathbf{M}} = 
\matrice{\bar{\inertia}_\elevation	&	\inertia_{\elevation\pitch} \\
		 \inertia_{\pitch\elevation}	&	\bar{\inertia}_\pitch	} 
, \quad
\bar{\mathbf{c}} = \matrice{ \bar{c}\dpitch^2 \\ \bar{c}\delevation^2},\\
\bar{\mathbf{g}} = \matrice{\frac{\mass_L}{2} l g \unitv^\perp \cdot \mathbf{e}_3 \\ 
		-\massR  l g \bar{\mathbf{R}}_B^W \offsetCoM \cdot \mathbf{e}_3 },\quad
\bar{\mathbf{Q}} = \matrice{0	&	0	\\
         -\offsetCoM\cdot\mathbf{e}_3 & 0 },			 
\end{gather*}
where
$\bar{\mathbf{R}}_B^W = \partial\mathbf{R}_B^W / \partial\pitch$,
$\bar{\inertia}_\elevation = \mass_L l^2/3$,
$\bar{\inertia}_\pitch = \massR\norm{\offsetCoM}^2$,
$\inertia_{\elevation\pitch} = \inertia_{\pitch\elevation} = -\massR l \bar{\mathbf{R}}_B^W \offsetCoM \cdot \unitv^\perp$, $\bar{c} = \massR l \bar{\mathbf{R}}_B^W \offsetCoM \cdot \unitv$.

\begin{figure}[t]
	\centering
	\captionsetup[subfigure]{farskip=0.5em,
							 captionskip=-1em,
							 labelformat=empty,
							 width=\columnwidth}
	\subfloat[][]{\includegraphics[width=\columnwidth]{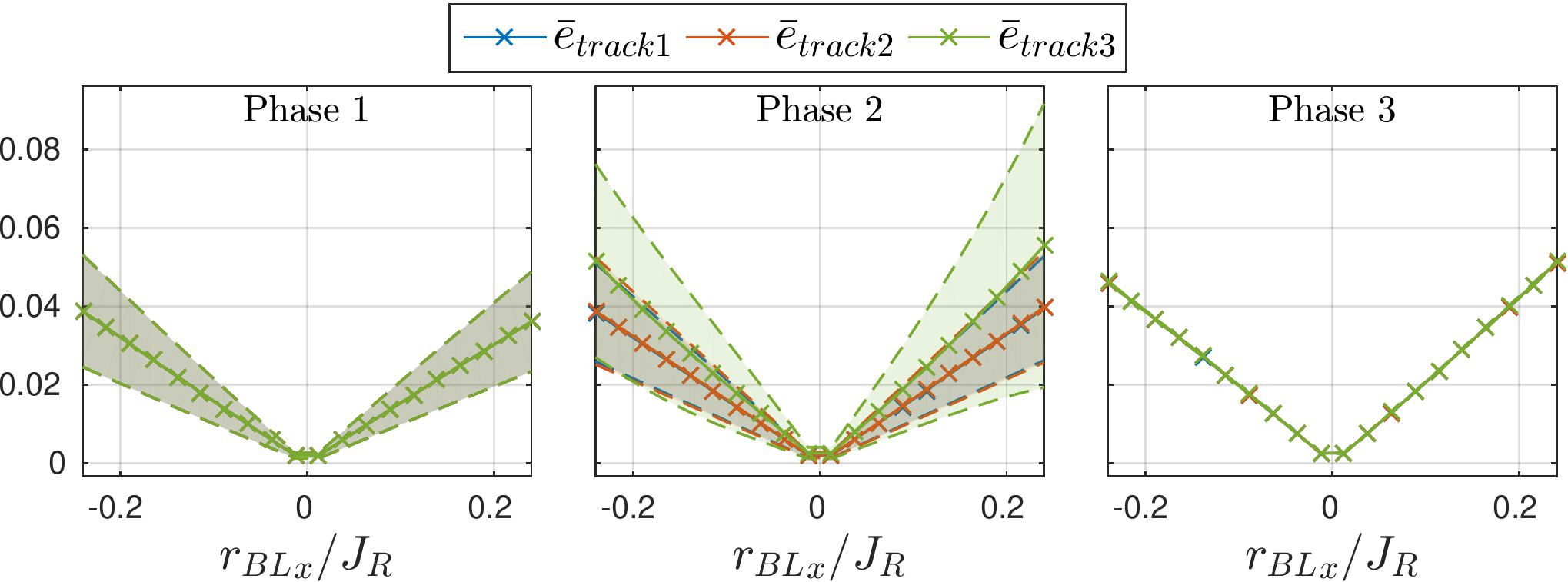}
	\label{fig:sim:generalModel:b:rBLx}}\\ 
	\subfloat[][]{\includegraphics[width=\columnwidth]{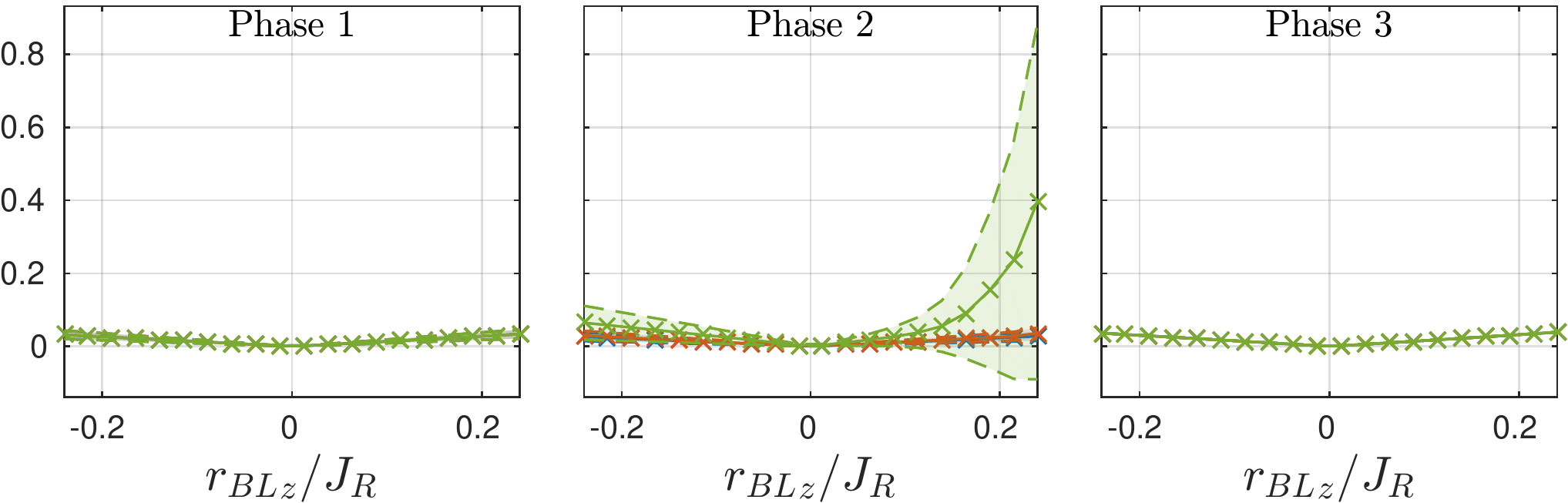}
	\label{fig:sim:generalModel:b:rBLz}}\\	
	\subfloat[][]{\includegraphics[width=\columnwidth]{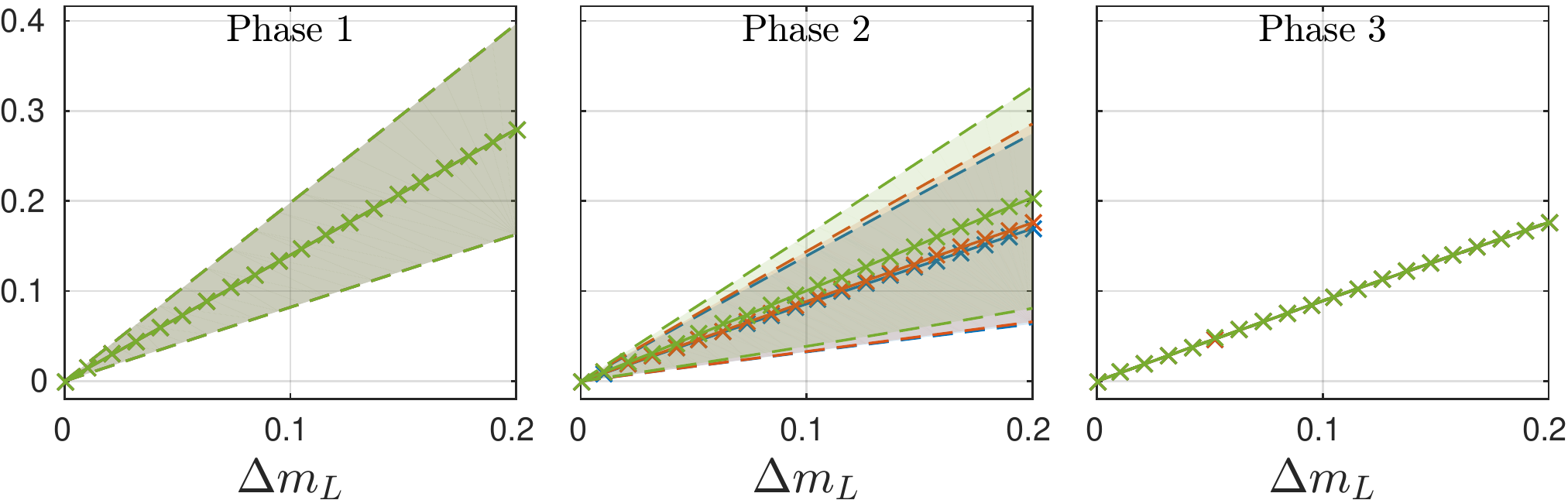}
	\label{fig:sim:generalModel:b:mL}}\\
	\caption{\emph{Controller $\Gamma^{b}$}: mean tracking error when changing the parameters of the general model.}
	\label{fig:sim:generalModel:b}
\end{figure}

For a plausible case in which the link consists of a cable of mass $\mass_L = 0.01\massR$ and inertia (during taut condition) $\inertia_L = \mass_L l^2/12$, and it is attached to the robot in the position $r_{BL} = [0.03 \;\, 0 \;\; 0.03]^T [\rm m]$ with respect to $\bFrame$, we noticed that the controlled system is stable but the error does not converge exactly to zero.
Indeed, due to the nonzero $\norm{\offsetCoM}$, the force along the link generates an extra torque on the aerial vehicle that is not compensated  and so a constant steady state error appears (the corresponding plots can be found in~\cite{2017-TogFra}).

In order to understand how each parameter of the more general model affects the tracking performances, as done in Sec.~\ref{subsec:paramVar}, we show in \fig\ref{fig:sim:generalModel:b} the mean tracking error and its standard deviation for different parameter values and in the three phases described before.
In particular the mass of the link is taken as $m_L = \Delta m_L \massR$.

We noticed that the negative effects due to a nonzero offset $\mathbf{r}_{BL}$ reduce or increase if the rotational inertia is increased or reduced, respectively.
Indeed, looking at the rotational dynamics in the case of non zero offset:
\begin{align}
{\ddpitch_R} = {\tau_R}/{\inertiaR} - \stress\unitv\cdot {\mathbf{r}_{BL}}/{\inertiaR},
\label{eqn:generalModel:attitudeDynamics}
\end{align}
one can notice that the effect of the link force on the angular acceleration decreases if the inertia increases.
Intuitively, a bigger inertia would mean a bigger mass or a bigger dimension of the vehicle. 
In the second case, the bigger the vehicle the more the effect of the offset becomes negligible.
For this reason in \fig\ref{fig:sim:generalModel:b} we plot the mean tracking error with respect to $r_{BL_x}/\inertiaR$ and $r_{BL_z}/\inertiaR$, thus normalizing this effect.

We did the same test for the controller $\Gamma^{a'}$, which resulted to be much more sensitive to link mass and to the offset than the controller $\Gamma^b$. This is due to the fact that one of the output, namely the attitude of the aerial vehicle $\pitch$, is directly influenced by the offset as it is shown in~\eqref{eqn:generalModel:attitudeDynamics}.
Furthermore, even with a small offset the tracking error is such that the actual trajectory passes through the singularity of the controller $\Gamma^{a'}$ (see \sect\ref{subsec:controller:a}) causing an unstable behavior. 

On the other hand, the controller $\Gamma^b$ turned to be much more robust to these sort of structural model variations. 
From \fig\ref{fig:sim:generalModel:b} we can see that the  parameters that mostly affect an increase of the error are the entries of $\mathbf{r}_{BL}$, i.e., ${r_{BL}}_x$ and ${r_{BL}}_z$.
One can notice that it is more advisable to attach the link such that one is sure that ${r_{BL}}_z \leq 0$, especially if agile motions are required.
The effect of the displacement along $\bx_R$ is instead almost symmetric. 
The small asymmetry is due to the particular trajectory passing from the first to the second quadrant.
The mean tracking error increases instead almost linearly with respect to the mass of the link. Nevertheless, even with $\mass_L$ equal to the $20\%$ of $\massR$  the closed loop system remains still perfectly stable.

\subsection{Noise on the Measurements}\label{subsec:noise}

\begin{figure}[t]
	\centering
	\captionsetup[subfigure]{width=\columnwidth}
	\subfloat[][Controller Results.]{\includegraphics[width=\columnwidth]{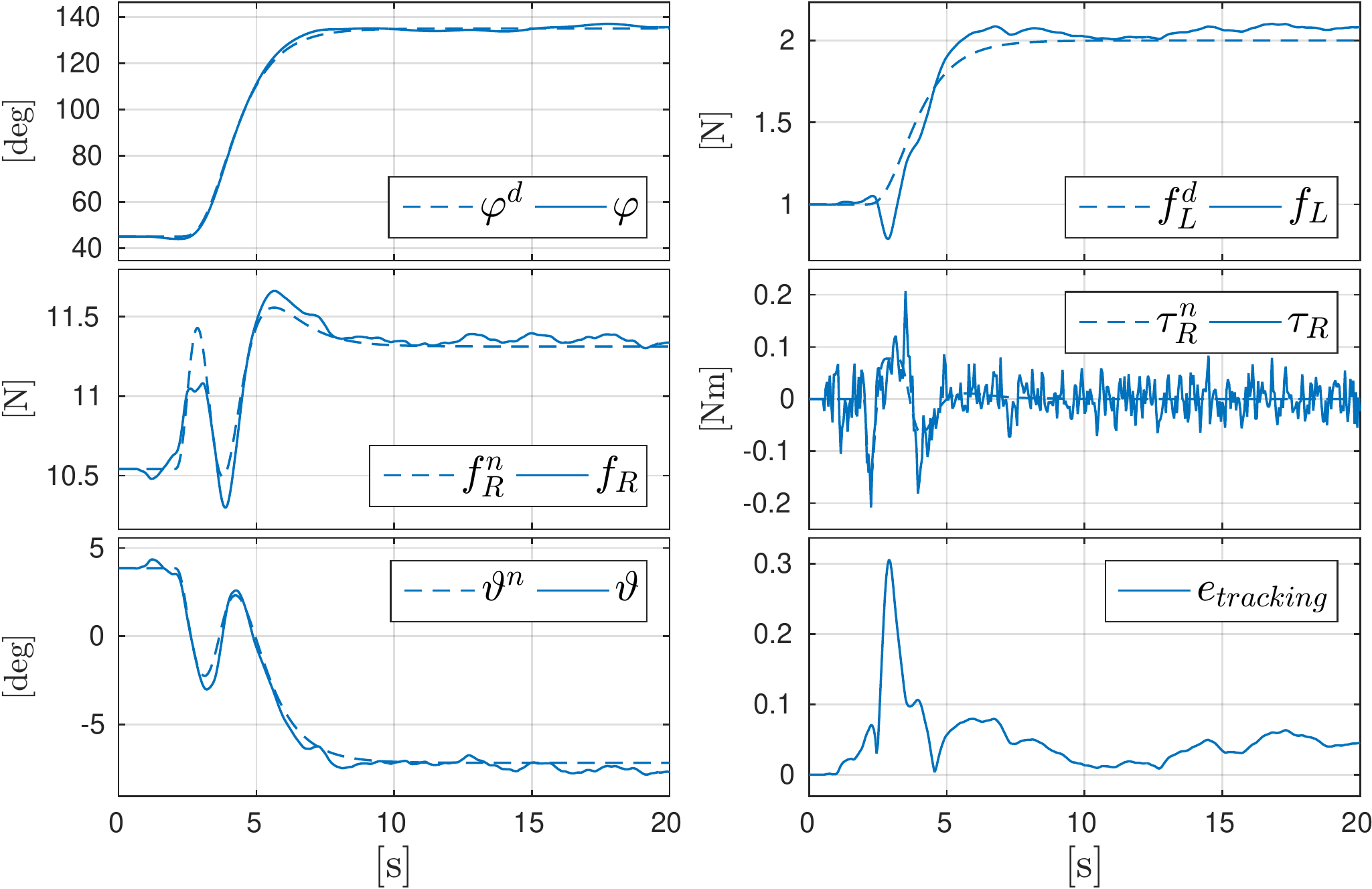}
	\label{fig:sim:noise:b:controller}}\\ 
	\subfloat[][Observer Results.]{\includegraphics[width=\columnwidth]{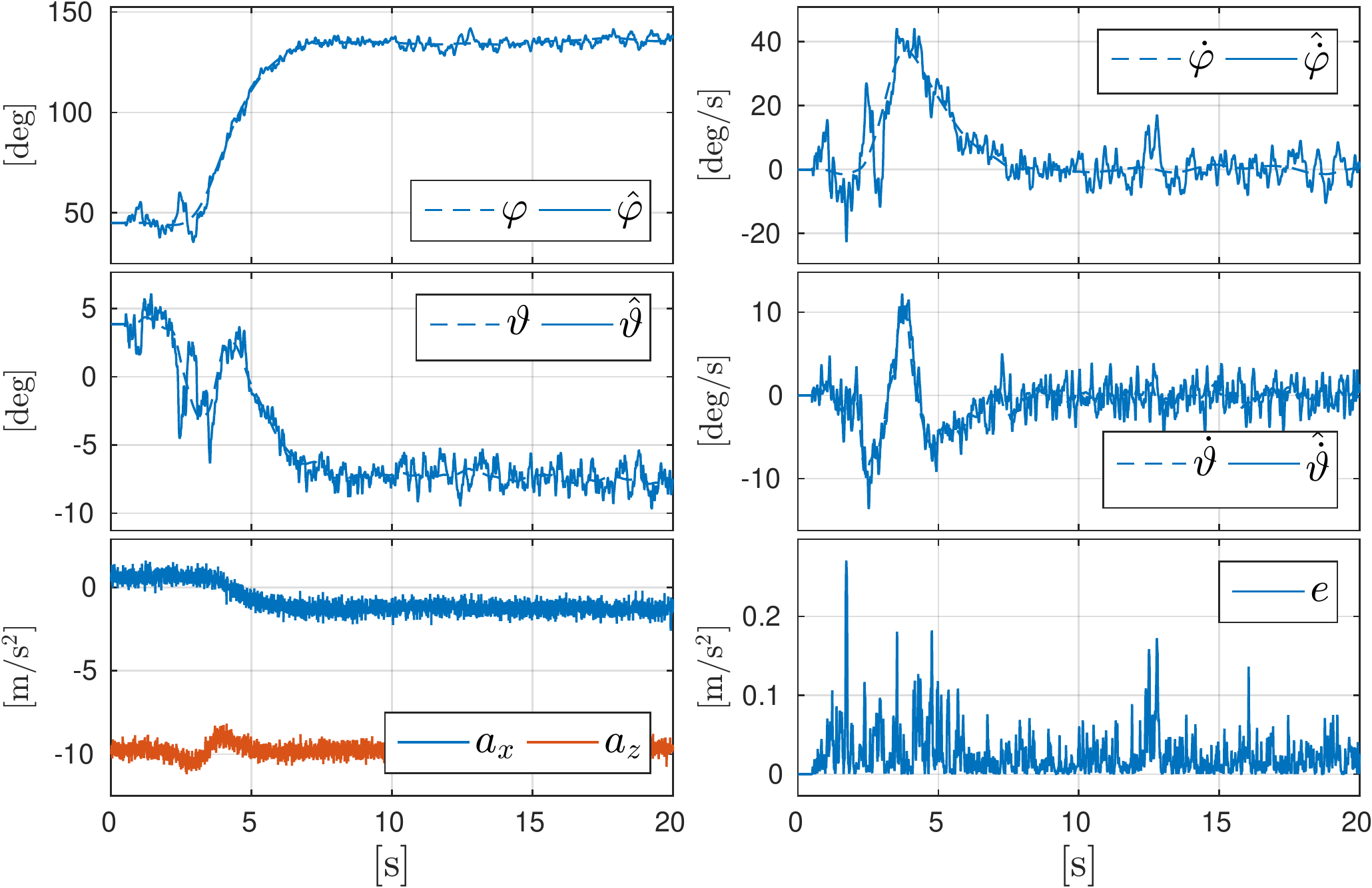}
	\label{fig:sim:noise:b:observer}}
	\small\caption{\emph{Controller $\Gamma^{b}$}: Noisy measurements. Similar kind of results have been observed also for the controller $\Gamma^{a'}$.}
	\label{fig:sim:noise:b}
\end{figure}

In this section we investigate the robustness of the proposed method in presence of noisy measurements, which always exist in reality.
We consider both the accelerometer and the gyroscope measures being affected by a white Gaussian noise of variance $0.1[\rm m/s^2]$ and $0.01[\rm rad/s]$ respectively.

From \fig\ref{fig:sim:noise:b} we can observe that the estimated state shows some noise but the corresponding error remains always bounded. 
Due to the noisy component on the estimated state the control action presents some oscillations that imply a non exact tracking of the desired trajectory.
Nevertheless the tracking error remains small and always bounded.
Similar results have been recorded also for the controller $\Gamma^{a'}$.
Notice that to achieve these results we had to reduce the gains of both the controller and observer. 
Indeed, high gain values increase the convergence speed but also amplify the sensitivity  to noisy measurements.
In general the two controllers does not show particularly different behaviors in face of the presence of noise.

\subsection{Non-ideal Motors}\label{subsec:timeConstant}
In a real scenario, one motor can not instantaneously change the spinning velocity of the propeller, and in turn the thrust produced. 
Instead the dynamics of the motor is characterized by a certain time constant, $\tau_M \in\nR{}$, that quantifies the time needed to change the motor speed.
We analyzed this additional non ideality testing the proposed method with different non-ideal motors characterized by an increasing time constant. 
The results, extensively presented and discussed in~\cite{2017-TogFra}, show that the tested system remains stable up to a time constant of $0.08[\rm s]$, which is completely acceptable in real systems.

\section{Conclusions}\label{sec:conclusion}

In this work we presented the main intrinsic characteristics of a tethered aerial robot system such as differential flatness, controllability and observability. 
We then used these results to design two different controllers, the first to achieve the tracking of elevation and pitch angles, and the second for the tracking of elevation angle and link force.
To close the control loop we designed a nonlinear observer able to retrieve the state from any dynamic condition using a inertial-only on-board sensors. This showed that inertial only measurements are enough to estimate the full state of the system.
An exhaustive numerical validation of the proposed method was presented considering typical non-ideal conditions such as parametric variation, more general model, noisy measurements and non-ideal motors.

We have recently started to build a real platform in order to implement and test the method under real experiments. Some preliminary experiments based on the theoretical discoveries presented here are already available in~\cite{2016i-TogTesRosFra}. In the future we plan to further extend such experimental system to take full advantage of the theory shown in this paper.
Another possible improvement would be to exploit the flatness property for motion planning respecting the limits and avoiding singularities.
Automatic identification of the system parameters and control adaptation are other possible directions of development.

\bibliographystyle{IEEEtran}
\bibliography{alias,main}

\begin{IEEEbiography}[{\includegraphics[width=1in,height=1.25in,clip,keepaspectratio]{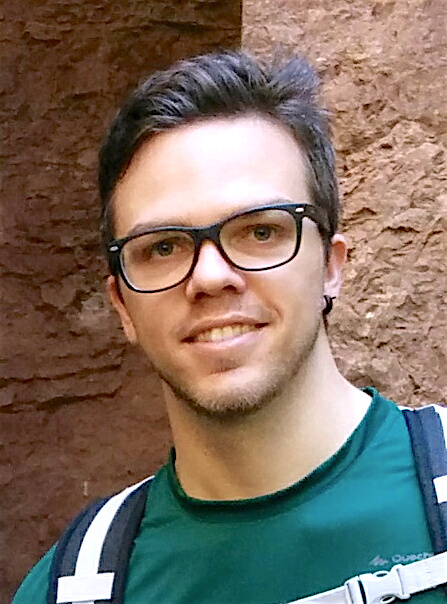}}]{Marco Tognon} (S'15) received the M.Sc. degree in automation engineering in 2014, from the University of Padua, Italy, with a master
thesis carried out at the Max Planck Institute for Biological Cybernetics, T\"bingen, Germany.
He is currently pursuing the Ph.D. degree at LAAS-CNRS, Toulouse, France.
His current research interests include robotics, unmanned aerial vehicles for physical interaction. 
\end{IEEEbiography}

\begin{IEEEbiography}[{\includegraphics[width=1in,height=1.25in,clip,keepaspectratio]{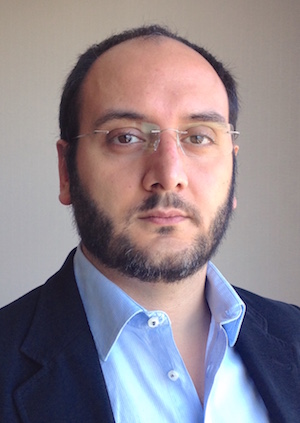}}]{Antonio Franchi} (S'07-M'11-SM'16) received the Ph.D.~degree in System Engineering from the Sapienza University of Rome, Italy, in 2010, and the Habilitation (HDR) in 2016 from the National Polytechnic Institute of Toulouse, France. In 2009 he was a Visiting Scholar at the University of California at Santa Barbara. From 2010 to 2014 he was first Research Scientist and then Senior Research Scientist and the Project Leader of the Autonomous Robotics and Human-Machine Systems group at the Max Planck Institute for Biological Cybernetics  in T\"ubingen, Germany. Since 2014 he has been a Permanent CNRS Researcher in the RIS team at LAAS-CNRS, Toulouse, France.
His main research interests are on autonomous systems, with a special regard to robot control and estimation, in particular for  multiple-robot systems and aerial robotics. He published more than 90 papers in peer-reviewed international journals and conferences.
In 2010 he was awarded  the IEEE RAS ICYA Best Paper Award.  He is an Associate Editor of the IEEE Transaction on Robotics. He is the co-founder of the IEEE RAS Technical Committee on Multiple Robot Systems and  of the International Symposium on Multi-robot and Multi-Agent Systems.
\end{IEEEbiography}

\end{document}